\documentclass[final]{cvpr}

\usepackage{times}
\usepackage{epsfig}
\usepackage{graphicx}
\usepackage{amsmath}
\usepackage{amssymb}
\usepackage[numbers,sort]{natbib}
\usepackage[utf8]{inputenc} % allow utf-8 input
\usepackage[T1]{fontenc}    % use 8-bit T1 fonts
\usepackage[colorlinks=true,linkcolor=red, citecolor=green]{hyperref}
\usepackage{url}            % simple URL typesetting
\usepackage{booktabs}       % professional-quality tables
\usepackage{amsfonts}       % blackboard math symbols
\usepackage{nicefrac}       % compact symbols for 1/2, etc.
\usepackage{microtype}      % microtypography
\usepackage{adjustbox}
\usepackage{multirow}
\usepackage[flushleft]{threeparttable}
\usepackage{makecell}
\usepackage{tabulary}
\usepackage{amsmath}
\usepackage{tabularx}
\usepackage{pifont}
\usepackage{kotex}
\DeclareMathOperator{\E}{\mathbb{E}}
\usepackage{bbold}
\usepackage{wrapfig}

\usepackage{enumitem}
\usepackage{graphicx}
\usepackage{scalerel}
\usepackage{caption}
\newcommand{\argmin}{\mathop{\mathrm{argmin}}}
\newcommand{\argmax}{\mathop{\mathrm{argmax}}}
\usepackage{stfloats}

\newcommand{\xmark}{\text{\ding{55}}}
\newcommand{\cmark}{\text{\ding{51}}}
\def\cvprPaperID{4014} % *** Enter the CVPR Paper ID here
\def\confYear{CVPR 2021}
\begin{document}

%%%%%%%%% TITLE
\title{BBAM: Bounding Box Attribution Map for Weakly Supervised \\Semantic and Instance Segmentation}

\author{Jungbeom Lee$^1$ ~~~~~~~ Jihun Yi$^1$ ~~~~~~~ Chaehun Shin$^1$  ~~~~~~~  Sungroh Yoon$^{1, 2, }$\thanks{Correspondence to: Sungroh Yoon <sryoon@snu.ac.kr>.}\\
$^1$ Department of Electrical and Computer Engineering, Seoul National University, Seoul, South Korea\\
$^2$ ASRI, INMC, ISRC, and Institute of Engineering Research, Seoul National University\\
{\tt\small \{jbeom.lee93, t080205, chaehuny, sryoon\}@snu.ac.kr}}

\maketitle

%%%%%%%%% ABSTRACT
\begin{abstract}
Weakly supervised segmentation methods using bounding box annotations focus on obtaining a pixel-level mask from each box containing an object. Existing methods typically depend on a class-agnostic mask generator, which operates on the low-level information intrinsic to an image. In this work, we utilize higher-level information from the behavior of a trained object detector, by seeking the smallest areas of the image from which the object detector produces almost the same result as it does from the whole image. These areas constitute a bounding-box attribution map (BBAM), which identifies the target object in its bounding box and thus serves as pseudo ground-truth for weakly supervised semantic and instance segmentation. This approach significantly outperforms recent comparable techniques on both the PASCAL VOC and MS COCO benchmarks in weakly supervised semantic and instance segmentation. In addition, we provide a detailed analysis of our method, offering deeper insight into the behavior of the BBAM.
The code is available at: \url{https://github.com/jbeomlee93/BBAM}.

\end{abstract}

% \vspace{-1em}
%%%%%%%%% BODY TEXT
\thispagestyle{empty}
\section{Introduction}
Object segmentation is one of the most important steps in image recognition.
Advances in deep learning have greatly improved the performance of semantic and instance segmentation~\cite{he2017mask, chen2017deeplab} through the use of huge amounts of pixel-level annotated training data. 
However, annotating with pixel-level masks requires a lot of effort. According to Bearman \textit{et al.}~\cite{bearman2016s}, constructing a pixel-level mask for an image containing an average of 2.8 objects takes about 4 minutes.
This is why weakly supervised methods have been proposed, in which segmentation networks are trained using annotations that are less detailed than pixel-level masks, such as bounding boxes~\cite{song2019box, khoreva2017simple, dai2015boxsup}, or image-level tags~\cite{lee2019ficklenet, ahn2019weakly, ahn2018learning}.

The most easily obtainable annotation is the class label.
Labeling an image with class labels takes around 20 seconds~\cite{bearman2016s},
but it only indicates that objects of certain classes are depicted and gives no information about their locations in the image. Moreover, class labels provide no help in separating different objects of the same class, which is the goal of instance segmentation.

Bounding boxes provide information about individual objects and their locations.
Bounding box annotation takes about 38.1 seconds per image~\cite{bellver2019budget}, which is much more attractive than constructing pixel-level masks.
Many researchers have tackled semantic segmentation~\cite{dai2015boxsup, khoreva2017simple, song2019box, kulharia12356box2seg} and instance segmentation~\cite{khoreva2017simple, liao2019weakly, sun2020weakly, hsu2019weakly, arun2020weakly} using bounding box annotations as a search space in which a class-agnostic object mask can be found by an off-the-shelf object mask generator. These are mostly based on GrabCut~\cite{rother2004grabcut} or multiscale combinatorial grouping (MCG)~\cite{pont2016multiscale}.
Those mask generators operate on the low-level information of images, such as the color or brightness of pixels, and this limits the quality of the resulting mask. 
Thus, applying these mask generators to bounding box annotations requires additional steps such as estimating what proportion of the pixels in a bounding-box belong to the corresponding object~\cite{song2019box, kulharia12356box2seg}, iterative refinement of an estimated mask~\cite{dai2015boxsup}, and auxiliary attention modules~\cite{kulharia12356box2seg}.

We propose a pixel-level method of localizing a target object inside its bounding box using a trained object detector.
We make use of attribution maps obtained from the trained object detector, which highlight the image regions that the detector focuses on in conducting object detection.
Inspired by the perturbation methods used to explain the output of image classifiers~\cite{fong2017interpretable, fong2019understanding, dabkowski2017real}, we introduce a bounding box attribution map (BBAM) which provides an indication of the smallest areas of an image that are sufficient to make an object detector produce almost the same result as that from the original image.
The BBAM identifies the area occupied by the object in each bounding box predicted by the trained object detector.
Since this localization takes place at the pixel level, it can be used as a pseudo ground truth for weakly supervised learning of semantic and instance segmentation.

The main contributions of this paper can be summarized as follows.
\begin{itemize}
\vspace{-4pt}
	\item[$\bullet$] We propose a bounding box attribution map (BBAM), which can draw on the rich semantics learned by an object detector to produce pseudo ground-truth for training semantic and instance segmentation networks.
	\vspace{-5pt}
	\item[$\bullet$] Our technique significantly outperforms previous state-of-the-art methods of weakly supervised semantic and instance segmentation, assessed on the PASCAL VOC 2012 and MS COCO 2017 benchmarks.
	\vspace{-5pt}
	\item[$\bullet$] We analyze our method from various viewpoints, providing deeper insights into the properties of the BBAM.
\end{itemize}

\section{Related Work}
Fully supervised semantic and instance segmentation based on pixel-level annotations is highly reliable, but the manual annotation process is laborious. This requirement is overcome by weakly supervised methods based on inexact, but easily obtainable, annotations such as scribbles~\cite{tang2018normalized}, bounding boxes~\cite{song2019box, khoreva2017simple}, or class labels~\cite{lee2019ficklenet, ahn2019weakly, sun2020mining}. In this section, we briefly review some recently introduced weakly supervised approaches that use class labels (Section~\ref{W_imagelabel}) or bounding boxes (Section~\ref{W_boxes}). In addition, we describe some visual saliency methods related to our method (Section~\ref{XAI}).

\subsection{Learning with Class Labels}\label{W_imagelabel}
A class activation map (CAM)~\cite{zhou2016learning} is a widely adopted technique to obtain a localization map from class labels. However, a CAM only identifies the most discriminative regions of objects~\cite{lee2019ficklenet, lee2019frame}, and hence the majority of existing methods that use class labels~\cite{lee2019frame, lee2019ficklenet, li2019guided, huang2018weakly, ahn2018learning, hou2018self, lee2018robust, Shimoda_2019_ICCV, JiangOAAICCV19, fan2018cian, hong2017weakly} are primarily concerned with expanding the area of the target object activated by a CAM. For instance, erasure methods~\cite{hou2018self, wei2017object} iteratively find new regions of the target object by removing discriminative regions in an image.
Other methods~\cite{fan2018cian, sun2020mining} consider the information shared between several images by capturing cross-image semantic similarities and differences.
Seed growing and refinement techniques~\cite{ahn2018learning, ahn2019weakly, huang2018weakly} are typically used to expand the regions representing the target object imperfectly that are in the initial CAM, on the basis of relationships between pixels.
Other methods construct CAMs that embody the multi-scale semantic context in an image~\cite{lee2019ficklenet, lee2018robust, wei2018revisiting}.
Despite these efforts, the information available from class labels remains limited, so auxiliary information acquired from web images~\cite{shen2018bootstrapping} or videos~\cite{hong2017weakly, lee2019frame} can be used together.

\subsection{Learning with Bounding Boxes}\label{W_boxes}
% \vspace{-0.2em}
Class labels have led to significant achievements in semantic segmentation, but they are inherently unhelpful in instance segmentation, which requires the separation of different objects of the same class.
In contrast, bounding boxes do provide information about the location of individual objects in an image, and they are still much cheaper than constructing pixel-level masks~\cite{bellver2019budget}.
Most existing methods utilized a bounding box as a search space to conduct low-level searches for object masks.
They create a pseudo mask within a box using off-the-shelf methods of mask proposal such as MCG~\cite{pont2016multiscale} or GrabCut~\cite{rother2004grabcut}. 
These processes can be guided by specifying the proportion of the pixels in a bounding box that are likely to belong to the object~\cite{song2019box, kulharia12356box2seg}. Iterative mask refinement techniques~\cite{dai2015boxsup} can also be applied.
However, these methods are largely based on low-level information in the image, and they ignore the semantics associated with the bounding boxes.
A rare exception is the multiple-instance learning formulation with a bounding box tightness prior~\cite{hsu2019weakly}: a crossing line within a box must contain at least one pixel of the target object.
The drawback with this approach is that only a small number of pixels are contributing to the localization of the object.

\vspace{-0.2em}
\subsection{Visual Saliency Methods}\label{XAI} 
\vspace{-0.2em}
Various methods have been proposed to visually explain the predictions of deep neural networks (DNNs)~\cite{fong2017interpretable, fong2019understanding, chang2018explaining, schulz2020restricting, zhou2016learning} in a form of a saliency map.
However, most studies have been concerned with classifiers, and only a few have looked at DNNs performing other tasks~\cite{hoyer2019grid, reif2019visualizing}. In particular, there have been no attempts to explain the predictions of object detectors, except Wu \textit{et al.}~\cite{wu2019towards}, who embedded interpretability inside the DNN, in this case Faster R-CNN~\cite{ren2015faster}. 
However, the explanation produced by their modified DNN is not immediately understandable because it is given as a form of tree, and thus it is not appropriate to generate pseudo ground truth for weakly supervised segmentation.
Gradient-based methods, such as SimpleGrad~\cite{zeiler2014visualizing}, SmoothGrad~\cite{smilkov2017smoothgrad}, and Grad-CAM~\cite{selvaraju2017grad}, can provide visual saliency maps of the results from classifiers, 
but these methods are not easily extended to object detectors, because of the structural difference between classifiers and object detectors. Nevertheless, gradient-based methods have a significant bearing on our approach, and we look at them in more detail in Section~\ref{sec_analysis}.

\begin{figure*}[t]
  \centering
  \includegraphics[width=0.92\linewidth]{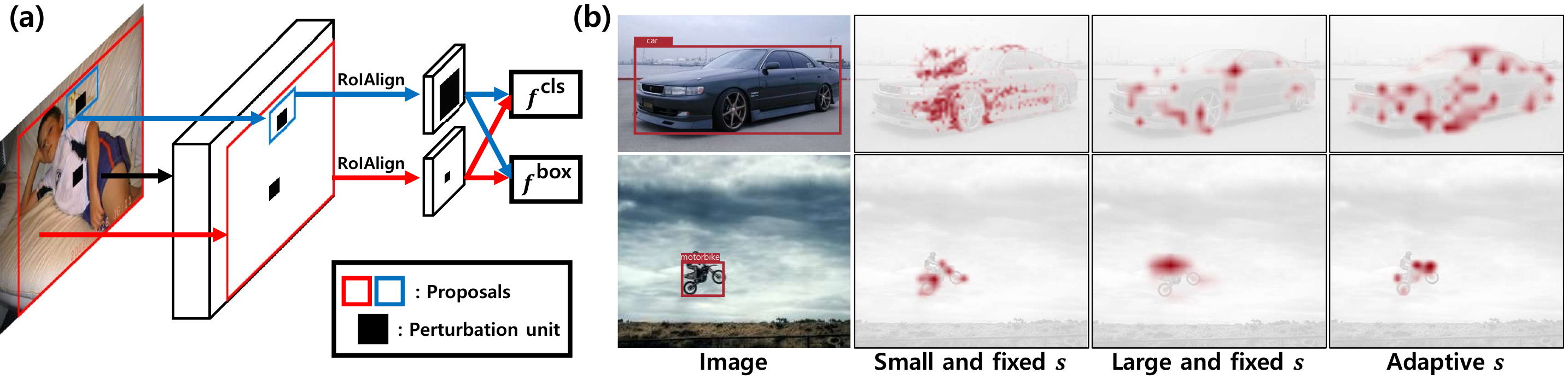} \\[-0.5em]
  \caption{\label{fig_ada_stride} The size of the \textit{perturbation unit} needs to be adjusted to the object size. (a) RoIAlign~\cite{he2017mask} produces perturbation units of different sizes. (b) Examples of resulting BBAMs with small fixed values of $s$, large fixed values of $s$, and values of $s$ determined adaptively. Fixed values of $s$, whether large or small, tend to generate unwanted artifacts.}
  \vspace{-0.8em}
\end{figure*}

\vspace{-0.2em}
\section{Method}
\vspace{-0.2em}
We first provide a brief description of the operation of object detectors in Section~\ref{revisit_FRCNN}. In Section~\ref{method_BBAM}, we introduce the BBAM for localizing objects in the bounding box. We then utilize the BBAM for weakly supervised semantic and instance segmentation in Sections~\ref{method_weakly} and~\ref{train_seg_net}.

% \vspace{-0.2em}
\subsection{Revisiting Object Detectors}\label{revisit_FRCNN}
% \vspace{-0.2em}
Modern object detectors can be divided into two categories: one-stage~\cite{liu2016ssd, lin2017focal, redmon2016you} and two-stage~\cite{ren2015faster, girshick2015fast} approaches.
We focus on two-stage object detectors such as Faster R-CNN~\cite{ren2015faster}, in which the two stages are region proposal and box refinement. 
A region proposal network (RPN) generates candidate object proposals in the form of bounding boxes; but these proposals are class-agnostic and noisy, and most of them are redundant, thereby necessitating a subsequent refinement step, in which classification and bounding box regression are performed on each proposal.
Since the proposal boxes proposed by the RPN are of different sizes, RoI pooling (\textit{e.g.,} RoIAlign~\cite{he2017mask}) is used to convert the feature map corresponding to each proposal to a predefined fixed size, as shown in Figure~\ref{fig_ada_stride}(a). The pooled feature map is then passed to the \textit{classification head} and also to the \textit{bounding box regression head}.

\textbf{Classification head.} It computes the class probability $p^c$ of class $c$ for each proposal and assigns the most likely class $c^{*} = \argmax_{c} p^c$ to the proposal.

\textbf{Bounding box regression head.} It adjusts the noisy proposal to fit the object by computing the offsets $t^c = (t^c_x, t^c_y, t^c_w, t^c_h)$ for each class $c \in \{1, 2, \cdots, C\}$.
The final localization is obtained by shifting each coordinate of the proposal using the offset $t^{c^{*}}$.
We refer to Ren \textit{et al.}~\cite{ren2015faster} for the details of the parameterization of each coordinate. 

For simplicity, we will abbreviate \textit{classification head} and \textit{bounding box regression head} as \textit{cls head} and \textit{box head}, respectively.

\subsection{Bounding Box Attribution Map}\label{method_BBAM}
Suppose we are given an image $I$ and the corresponding bounding box annotations.
We also have a set of object proposals $\mathcal{O} = \{o_k\}^{K}_{k=1}$, either given or obtained by RPN, where $K$ is the number of proposals. For each proposal $o_k$, the \textit{box head} $f^{\text{box}}$ and the \textit{cls head} $f^{\text{cls}}$ produce box offsets $t_k = f^{\text{box}}(I, o_k)$ and the class probability $p_k = f^{\text{cls}}(I, o_k)$, respectively. We omit the proposal indices $k$ for brevity.

The bounding box attribution map (BBAM) identifies the important region in the image that the detector needs to perform object detection.
We find the smallest mask \mbox{$\mathcal{M}: \Omega \rightarrow [0, 1]$} where $\Omega$ is a set of pixels, which captures a subset of the image that produces almost the same prediction as the original image.
A small $\mathcal{M}$ reduces the amount of unnecessary information reaching the detector.
The mask specifies a subset of the image in terms of the perturbation function $\Phi(I, \mathcal{M}) = \!I \circ \mathcal{M} \! + \! \mu \circ (1 \!-\! \mathcal{M})$, where $\circ$ denotes pixel-wise multiplication, and $\mu$ is the per-channel mean of the training data with the same size as $\mathcal{M}$.
For each proposal $o$, the best mask $\mathcal{M}^{*}$ is obtained by optimizing the following function using gradient descent with respect to $\mathcal{M}$:
\begin{align}\label{eq_mask}
\mathcal{M}^{*} = \argmin_{\mathcal{M} \in [0, 1]^{\Omega}} \lambda \left\lVert \mathcal{M} \right\rVert_1 + \mathcal{L}_\text{perturb},
\end{align}
\vspace{-1em}
\begin{align}\label{eq_mask_perturb}
\begin{split}
\mathcal{L}_\text{perturb} = & \hspace{5pt}\mathbb{1}_{\text{box}} \left\lVert t^c - f^{\text{box}}(\Phi(I, \mathcal{M}), o) \right\rVert_1\\ &+ \mathbb{1}_{\text{cls}} \left\lVert p^c - f^{\text{cls}}(\Phi(I, \mathcal{M}), o) \right\rVert_1,
\end{split}
\end{align}
where $\mathbb{1}_{\text{box}}$ and $\mathbb{1}_{\text{cls}}$ are logical variables that have a value of 0 or 1, to control which head is used to produce localizations,
and $t^c = f^{\text{box}}(I, o)$ and $p^c = f^{\text{cls}}(I, o)$ are the predictions for the original image.

Previous studies show that using a mask of the same spatial size as the input image incurs undesirable artifacts due to the adversarial effect~\cite{goodfellow2014explaining}: even a perturbation in a tiny magnitude can significantly change the prediction of a DNN.
This problem can be addressed by introducing a coarse mask downsampled by a stride $s$~\cite{fong2017interpretable, fong2019understanding, dabkowski2017real, hoyer2019grid}, so multiple image pixels are perturbed by a single element of $\mathcal{M}$. 
We can then optimize $\mathcal{M} \in \mathbb{R}^{\lceil w/s \rceil \times \lceil h/s \rceil}$ for the image $I \in \mathbb{R}^{w \times h}$, using the perturbation function $\Phi(I, \mathcal{M}) = I \circ \hat{\mathcal{M}} + \mu \circ (1 - \hat{\mathcal{M}})$, where $\hat{\mathcal{M}} \in \mathbb{R}^{w \times h}$ is upsampled $\mathcal{M}$ to a width of $w$ pixels and a height of $h$ pixels.

Existing methods of explaining the output of classifiers~\cite{fong2017interpretable, fong2019understanding, dabkowski2017real} or semantic segmentation networks~\cite{hoyer2019grid} use a fixed value of $s$ for all images, \textit{i.e.}, they fix the size of a \textit{perturbation unit}\footnote{The \textit{perturbation unit} is a block of image pixels perturbed by a single element of $\mathcal{M}$.}.
However, in the case of object detectors, a \textit{perturbation unit} of fixed size can result in perturbations of different sizes to the RoI-pooled features, depending on the size of the proposals, as shown in Figure~\ref{fig_ada_stride}(a).
Figure~\ref{fig_ada_stride}(b) shows how the size of a \textit{perturbation unit}, after RoI pooling, can fail to match the sizes of target objects: the perturbations are too coarse for small objects and too fine for large objects.
Therefore, we use an adaptive stride $s(a)$ where $a$ is the ratio of the area of the bounding box predicted by the object detector to that of the image, so that we use a small stride for a small object and a large stride for a large object.

\subsection{Generating Pseudo Ground Truth}\label{method_weakly}
Since the BBAM is a pixel-level localization of the target object in a bounding box predicted by the object detector, it can be used as pseudo ground-truth for weakly supervised semantic and instance segmentation, using the following procedure:
We first train an object detector, then create pseudo ground-truth semantic and instance masks for training images, using the BBAM of the trained object detector. These pseudo ground-truth masks can then be used to train semantic and instance segmentation networks. We will now explain this procedure in more detail.

\textbf{Creating masks.} 
Multiple proposals on a single object yield multiple predictions from the object detector. 
In order to benefit from the diversity of these predictions, we build the pseudo ground-truth from the BBAMs of multiple proposals.
For each ground-truth box, we generate a set of object proposals $\mathcal{O}$ by randomly jittering each coordinate of the box by up to $\pm30\%$.
These proposals are sent to the $f^{\text{cls}}$ and the $f^{\text{box}}$. If the $f^{\text{cls}}$ correctly predicts the ground-truth class, and the intersection over union (IoU) value associated with the predicted box by $f^{\text{box}}$ is greater than 0.8, then the proposal is added to a set of positive proposals $\mathcal{O}^+ \subset \mathcal{O}$.
We then use a modified version of $\mathcal{L}_{\text{perturb}}$ in Eq.~\ref{eq_mask} to amalgamate all the positive proposals into a single localization map, as follows:
\begin{align}\label{eq_mask_weak}
\begin{split}
\mathcal{L}_{\text{perturb}} = \E_{o\in \mathcal{O}^{+}}[&\mathbb{1}_{\text{box}} \left\lVert t^c - f^{\text{box}}(\Phi(I, \mathcal{M}), o) \right\rVert_1\\ &+ \mathbb{1}_{\text{cls}} \left\lVert p^c - f^{\text{cls}}(\Phi(I, \mathcal{M}), o) \right\rVert_1].
\end{split}
\end{align}
In this equation both $\mathbb{1}_{\text{box}}$ and $\mathbb{1}_{\text{cls}}$ are set to 1, since the BBAMs of $f^{\text{box}}$ and $f^{\text{cls}}$ provide complementary localization results (see Section~\ref{sec_analysis} for details).
A BBAM obtained in this way may partially cover the target object because not all pixels of the object are considered by $f^{\text{box}}$ and $f^{\text{cls}}$. 
Therefore we refine the BBAM using CRFs~\cite{krahenbuhl2011efficient}, following previous work~\cite{song2019box, ahn2018learning, khoreva2017simple}.
Finally, we create pseudo instance-level ground-truth masks by considering the pixels in each BBAM with values greater than a threshold $\theta$ to be foreground.
We denote such a mask as $\mathcal{T}$.

The threshold $\theta$ controls the size of $\mathcal{T}$.
However, the proportion of pixels in each BBAM which correspond to the foreground will vary, so it may not be appropriate to use a fixed $\theta$. Therefore we introduce two thresholds $\theta_{\text{fg}}$ and $\theta_{\text{bg}}$: pixels whose attribution values are higher than $\theta_{\text{fg}}$ are considered to be part of the foreground, and pixels whose values are lower than $\theta_{\text{bg}}$ are considered to be part of the background. The remaining pixels are ignored in the loss computations during training segmentation networks.

\textbf{Refine with MCG proposals.} 
MCG~\cite{pont2016multiscale} is an unsupervised mask proposal generator, which is commonly used in weakly supervised instance segmentation~\cite{zhou2018weakly, zhu2019learning, arun2020weakly, liu2020leveraging, khoreva2017simple}.
We can use mask proposals generated by MCG to refine a mask $\mathcal{T}$.
We first select the mask proposal that has the highest IoU with $\mathcal{T}$.
However, that proposal may partially cover the target object. We therefore consider other proposals that are completely contained within $\mathcal{T}$.
More formally, given a set of MCG proposals $\{m_i\}_{i=1}^K$, the refined mask $\mathcal{T}_{r}$ is derived as follows:
\begin{align}\label{eq_mcg}
\begin{split}
&\mathcal{T}_{r}= \bigcup\limits_{i\in \mathcal{S}} m_{i} \text{,~~~where} \\
&\mathcal{S} = \{i~|m_i \subset \mathcal{T}\} \cup \{\argmax_i \text{IoU}(m_i, \mathcal{T})\}.
\end{split}
\end{align}

\subsection{Training the Segmentation Network}\label{train_seg_net}
We now explain the procedure that we use for training the semantic and instance segmentation network.

\textbf{Instance segmentation.}
We use Mask R-CNN~\cite{he2017mask}, pre-trained on ImageNet~\cite{deng2009imagenet}.
We use a seed growing technique~\cite{huang2018weakly, ahn2018learning, lee2019ficklenet, lee2019frame} for pseudo-labeling the pixels ignored during training: 
Starting with the pixels identified by the initial pseudo ground-truth mask, more of the ignored pixels progressively participate in the loss computation as training proceeds. We refer to Huang \textit{et al.}~\cite{huang2018weakly} for more details.

\textbf{Semantic segmentation.}
We use DeepLab-v2~\cite{chen2017deeplab}, pre-trained on the ImageNet~\cite{deng2009imagenet} dataset. The pseudo labels produced in Section~\ref{method_weakly} can easily be made suitable for semantic segmentation by converting them from instance-level to class-level.
Pixels assigned to two or more object classes are ignored during the loss computation.

\section{Experiments}
% \vspace{-0.3em}
\subsection{Experimental Setup}\label{setup}
% \vspace{-0.3em}
\textbf{Dataset and evaluation metrics.} We conducted experiments on the PASCAL VOC~\cite{everingham2010pascal} and the MS COCO datasets~\cite{lin2014microsoft}.
The PASCAL VOC dataset contains 20 object classes and one background class. 
Following the same protocol as other recent work on weakly supervised semantic and instance segmentation~\cite{ahn2019weakly, arun2020weakly, hsu2019weakly, song2019box}, we used an augmented set of 10,582 training images produced by Hariharan \textit{et al.}~\cite{hariharan2011semantic}. 
The MS COCO dataset has 118K training images containing 80 object classes.
We report mean intersection-over-union (mIoU) values for semantic segmentation. For instance segmentation, we report average precision (AP$_{\tau}$) at IoU thresholds $\tau$; averaged AP over IoU thresholds from 0.5 to 0.95; and the average best overlap (ABO). 

\textbf{Reproducibility.}
We used the PyTorch~\cite{paszke2017automatic} implementation~\cite{massa2018mrcnn} of Faster R-CNN~\cite{ren2015faster} and Mask R-CNN~\cite{he2017mask}.
For semantic segmentation, we used the PyTorch implementation of DeepLab-v2-ResNet101~\cite{pytorchdeeplab}.
We set $s(a)$ to $16+48\sqrt{a}$ and $\lambda$ to $0.007$. We set $\theta_{\text{fg}}$ and $\theta_{\text{bg}}$ to 0.8 and 0.2 respectively. To find $\mathcal{M}^{*}$ in Eq.~\ref{eq_mask}, we used Adam optimizer~\cite{kingma2014adam} with a learning rate of 0.02 for 300 iterations. The experiments were performed on NVIDIA Tesla V100 GPUs. For MCG mask proposals, we used the pre-computed proposals for PASCAL VOC and MS COCO images provided by Pont-Tuset \textit{et al.}~\cite{pont2016multiscale}.

\begin{table}[t]
\renewcommand{\arraystretch}{0.97}
\centering
\caption{Weakly supervised instance segmentation performance on PASCAL VOC 2012 \textit{val} images.}
\vspace{-0.8em}
  \label{instance_voc}
  
  \begin{tabular}{l @{\hskip 0.04in} c@{\hskip 0.04in}c @{\hskip 0.04in} c @{\hskip 0.04in}c@{\hskip 0.04in}c}
    \Xhline{1pt}\\[-0.95em]
     Method   & AP$_{25}$    & AP$_{50}$   & AP$_{70}$ & AP$_{75}$ & ABO\\
\hline\hline
    \\[-0.9em]
    \multicolumn{4}{l}{Full supervision: Instance masks } \\
    $\text{MNC}_{\text{~~CVPR '16}}$~\cite{dai2016instance} & - & 63.5 & 41.5 & -  & -  \\
    $\text{Mask R-CNN}_{\text{~~ICCV '17}}$~\cite{he2017mask}  
    & 77.3&   69.1   & 49.9    & 41.9  &  65.8 \\
  
    \\[-0.9em]
    \hline
    \\[-0.9em]
    \multicolumn{4}{l}{Weak supervision: Image-level tags } \\

    $\text{PRM}_{\text{~~CVPR '18}}$~\cite{zhou2018weakly} 
    & 44.3 &   26.8   &  -   & 9.0 & 37.6 \\
    $\text{IAM}_{\text{~~CVPR '19}}$~\cite{zhu2019learning}   
    & 45.9 & 28.8    &  -    & 11.9 & 41.9\\
    $\text{Label-PEnet}_{\text{~~ICCV '19}}$~\cite{ge2019label} 
    & 49.1 &   30.2 & - &      12.9 & 41.4 \\
    $\text{CountSeg}_{\text{~~CVPR '19}}$~\cite{cholakkal2019object} 
    & 48.5 & 30.2 &  -   &  14.4 & 44.3\\
    $\text{IRNet}_{\text{~~CVPR '19}}$~\cite{ahn2019weakly}   
    & - & 46.7    &  23.5    & -  & - \\
    $\text{Kim \textit{et al.}}_{\text{~~WACV '21}}$~\cite{hwang2021weakly} 
    & 56.6 & 38.1    &  -   & 12.3  & 48.2 \\
    $\text{LIID}_{\text{~~TPAMI '20}}$~\cite{liu2020leveraging}
    & - & 48.4 &  -  &  24.9 & 50.8 \\
    $\text{Arun \textit{et al.}}_{\text{~~ECCV '20}}$~\cite{arun2020weakly}  
    & 59.1 & 49.7 & 29.2  &  27.1 & - \\
    \\[-0.9em]
    \hline
    \\[-0.9em]
    \multicolumn{4}{l}{Weak supervision: Bounding boxes}\\
    $\text{SDI}_{\text{~~CVPR '17}}$~\cite{khoreva2017simple} 
    & - & 44.8 &  -   & 16.3 & 49.1 \\
    $\text{Liao \textit{et al.}}_{\text{~~ICASSP '19}}$~\cite{liao2019weakly} 
    & -& 51.3 &    -  &  22.4 & 51.9 \\
    $\text{Sun \textit{et al.}}_{\text{~~Access '20}}$~\cite{sun2020weakly}  
    &- & 56.9 &    -  &  21.4 & 56.9 \\
    
    $\text{Hsu \textit{et al.}}_{\text{~~NeurIPS '19}}$~\cite{hsu2019weakly} 
    & 75.0 & 58.9 &    30.4  &  21.6 & - \\
    $\text{Arun \textit{et al.}}_{\text{~~ECCV '20}}$~\cite{arun2020weakly} 
    & 73.1 & 57.7 &    33.5  &  31.2 & - \\
   
    BBAM (Ours) 
    & \textbf{76.8} & \textbf{63.7} &    \textbf{39.5}  &  \textbf{31.8} & \textbf{63.0} \\
    % \\[-0.9em]
    \Xhline{1pt}
    \vspace{-2.1em}
    \end{tabular}%
    
      \end{table}

\subsection{Weakly Supervised Instance Segmentation}

\begin{figure*}[t]
  \centering
  \includegraphics[width=0.95\linewidth]{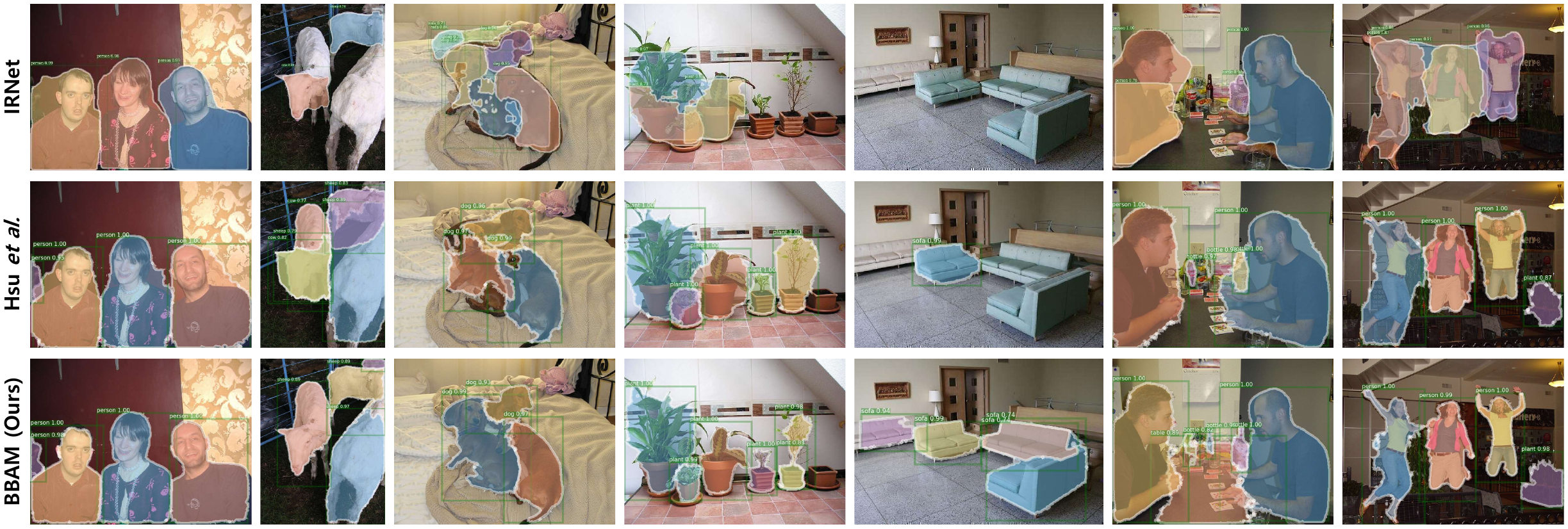} \\[-0.8em]
  \caption{\label{ins_sample} Examples of predicted instance masks for PASCAL VOC \textit{val} images of IRNet~\cite{ahn2019weakly}, Hsu \textit{et al.}~\cite{hsu2019weakly}, and ours.}
  \vspace{-1em}
\end{figure*}

\textbf{Results on PASCAL VOC.} Table~\ref{instance_voc} compares the performance of our method with that of other recent methods of weakly supervised instance segmentation which use image-level tags or bounding boxes. Our method significantly outperforms those methods. 
Specifically, the AP$_{50}$ and AP$_{70}$ values of our method are both 6.0\% higher than those of the previous best performing method which also uses bounding box annotation~\cite{arun2020weakly}.
We include results from two fully supervised methods: MNC~\cite{dai2016instance} and Mask R-CNN~\cite{he2017mask}. 
The performance of Mask R-CNN~\cite{he2017mask}, which is fully supervised, can be viewed as an upper bound on the achievable performance of our method.
We achieve 92.2\% and 95.7\% of the performance of fully supervised Mask R-CNN, in terms of AP$_{50}$ and ABO respectively. Figure~\ref{ins_sample} presents examples of instance masks produced by our method.

\textbf{Results on MS COCO 2017.}
This is a challenging dataset containing more objects in an image on average than PASCAL VOC.
The sizes of instances of objects are also more diverse.
Table~\ref{instance_coco} compares the performance of our method with that of other weakly supervised instance segmentation methods with various levels of supervision on MS COCO. Our method achieves a 6.7\% higher value of AP$_{75}$ than the previous best performing method which uses bounding box annotations. Since the labels for \textit{test-dev} images are not publicly available, the results for the \textit{test-dev} images were obtained from the MS COCO challenge website.

\begin{table}[t]
\renewcommand{\arraystretch}{0.97}
\centering
\caption{Comparison of instance segmentation methods with various types of supervision on MS COCO. The results of Hsu \textit{et al.}~\cite{hsu2019weakly} were obtained from \href{https://media.nips.cc/nipsbooks/nipspapers/paper_files/nips32/reviews/3565-AuthorFeedback.pdf}{here}.}
\vspace{-0.8em}
  \label{instance_coco}
  \begin{threeparttable}
  \begin{tabular}{l @{\hskip 0.08in} c@{\hskip 0.12in}c @{\hskip 0.12in} c @{\hskip 0.1in}c}
    \Xhline{1pt}\\[-0.95em]
     Method   & sup.    & AP   & AP$_{50}$ & AP$_{75}$ \\
\hline\hline
    \\[-0.9em]
    
    \multicolumn{4}{l}{MS COCO \textit{val} images} \\

    $\text{Mask R-CNN}_{\text{~~ICCV '17}}$~\cite{he2017mask} & $\mathcal{F}$ & 35.4  & 57.3  & 37.5\\
    $\text{Shen \textit{et al.}}_{\text{~~CVPR '19}}$~\cite{shen2019cyclic}  & $\mathcal{I}$ & 6.1   & 11.7  & 5.5 \\
    $\text{Laradji \textit{et al.}}_{\text{~~arXiv '19}}$~\cite{laradji2019instance} & $\mathcal{I}, \mathcal{P}$ &      7.8 &     18.2  &   8.8 \\
    $\text{Hsu \textit{et al.}}_{\text{~~NeurIPS '19}}$~\cite{hsu2019weakly} & $\mathcal{B}$& 21.1  & 45.5    & 17.2 \\
    BBAM (Ours)  & $\mathcal{B}$ &     \textbf{26.0}  &  \textbf{50.0}     &   \textbf{23.9} \\
    \\[-0.9em]
    \hline
    \\[-0.9em]
    \multicolumn{4}{l}{MS COCO \textit{test-dev} images}\\
    $\text{Mask R-CNN}_{\text{~~ICCV '17}}$~\cite{he2017mask} & $\mathcal{F}$  &  35.7     &   58.0    & 37.8 \\
    $\text{Fan \textit{et al.}}_{\text{~~ECCV '18}}$~\cite{fan2018associating}  & $\mathcal{I}, \mathcal{S}_I$  & 13.7  & 25.5  & 13.5 \\
    $\text{LIID}_{\text{~~TPAMI '20}}$~\cite{liu2020leveraging}  & $\mathcal{I}$   & 16.0  & 27.1    & 16.5 \\
   BBAM (Ours)  & $\mathcal{B}$ &     \textbf{25.7}  &  \textbf{50.0}     &   \textbf{23.3} \\
    % \\[-0.9em]
    \Xhline{1pt}
    % \vspace{-2em}
    \end{tabular}%
    % \vspace{-0.1em}
    \begin{tablenotes}
  \footnotesize
\item $\mathcal{F}-$Full, $\mathcal{I}-$Image label, $\mathcal{P}-$Point, $\mathcal{B}-$Box, $\mathcal{S}_I-$Instance saliency
        \end{tablenotes}
     \end{threeparttable}
    \vspace{-0.7em}
      \end{table}
\begin{table}[t]
\renewcommand{\arraystretch}{0.97}
\centering
  \caption{Weakly supervised semantic segmentation on PASCAL VOC 2012 \textit{val} and \textit{test} images.}  \label{table_semantic}
\vspace{-0.8em}
\begin{tabular}{l@{\hskip 1.in}c@{\hskip 0.2in}c}
    \Xhline{1pt}\\[-0.95em]
    Method  & \textit{val} & \textit{test}\\
    \hline\hline 
    \\[-0.9em]
    
    \multicolumn{3}{l}{Full supervision: Semantic masks} \\
    $\text{DeepLab}_{\text{~~TPAMI '17}}$~\cite{chen2017deeplab}    & 76.8  & 76.2 \\
    \\[-0.9em]
\hline
    \\[-0.9em]
    \multicolumn{3}{l}{Weak supervision: Image-level tags}\\

    $\text{FickleNet}_{\text{~~CVPR '19}}$~\cite{lee2019ficklenet}   & 64.9 & 65.3\\
    $\text{CIAN}_{\text{~~AAAI '20}}$~\cite{fan2018cian}     & 64.3  & 65.3 \\
    $\text{Chang \textit{et al.}}_{\text{~~CVPR '20}}$~\cite{chang2020weakly}     & 66.1  & 65.9\\
    $\text{Sun \textit{et al.}}_{\text{~~ECCV '20}}$~\cite{sun2020mining}     & 66.2  & 66.9  \\

    \\[-0.9em]
    \hline
    \\[-0.9em]
\multicolumn{3}{l}{Weak Supervision: Bounding boxes}\\
$\text{WSSL}_{\text{~~ICCV '15}}$~\cite{Papandreou_2015_ICCV}  & 60.6   & 62.2 \\
$\text{BoxSup}_{\text{~~ICCV '15}}$~\cite{dai2015boxsup}    & 62.0    & 64.6  \\
    $\text{SDI}_{\text{~~CVPR '17}}$~\cite{khoreva2017simple}    & 69.4  & -  \\
    $\text{Song \textit{et al.}}_{\text{~~CVPR '19}}$~\cite{song2019box}    & 70.2  & - \\
    BBAM (Ours) & \textbf{73.7} & \textbf{73.7}  \\
    \Xhline{1pt}
    \end{tabular}%
    
    % \end{adjustbox}
    \vspace{-0.6em}

      \end{table}

\begin{figure*}[t]
  \centering
  \includegraphics[width=0.95\linewidth]{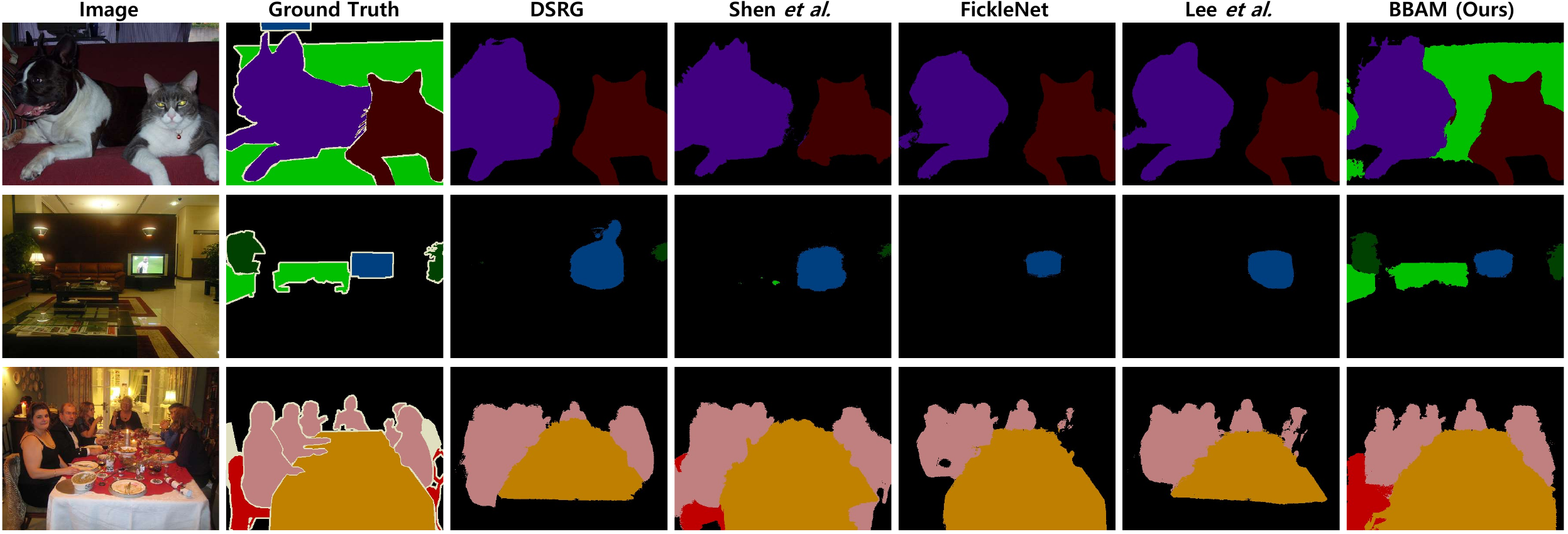} \\[-0.7em]
  \caption{\label{seg_sample} Examples of predicted semantic masks for PASCAL VOC \textit{val} images of DSRG~\cite{huang2018weakly}, Shen \textit{et al.}~\cite{shen2018bootstrapping}, FickleNet~\cite{lee2019ficklenet}, Lee \textit{et al.}~\cite{lee2019frame}, and our method.}
  \vspace{-1.4em}
\end{figure*}

\subsection{Weakly Supervised Semantic Segmentation}
Table~\ref{table_semantic} compares published mIoU values achieved by recent methods performing semantic segmentation on validation and test images from the PASCAL VOC 2012 dataset. Since the labels for test images are not publicly available, the results for the test images were obtained from the official PASCAL VOC evaluation server.
Our method, using the BBAM, yields an mIoU value of 73.7 for both the validation and the test images in the PASCAL VOC 2012 semantic segmentation benchmark. Our method outperforms all the methods that use image-level tags or bounding boxes for supervision. 
This new state-of-the-art performance was achieved with vanilla DeepLab-v2~\cite{chen2017deeplab} without any modifications to networks or additional training techniques, such as label refinement during training~\cite{dai2015boxsup}, recursive training~\cite{khoreva2017simple}, or fine-tuning with additional losses~\cite{song2019box}. Figure~\ref{seg_sample} presents examples of semantic masks produced by our method.

% Table generated by Excel2LaTeX from sheet 'Sheet1'
\begin{table}[t]
\renewcommand{\arraystretch}{0.92}
  \centering
  \caption{Effectiveness of using MCG proposals for instance segmentation. AP$_{S}$, AP$_{M}$, and AP$_{L}$ respectively denote the AP values for small, medium, and large objects. }
  \vspace{-1em}
    \begin{tabular}{ccccccc}
    \Xhline{1pt}\\[-0.95em]
    
    MCG  & AP   & AP$_{50}$ & AP$_{75}$  & AP$_{S}$   & AP$_{M}$   & AP$_{L}$ \\
    \hline\hline
    \\[-0.9em]  
    \multicolumn{7}{l}{PASCAL VOC \textit{val} images:} \\
     \xmark     &     29.6  &   61.9    &   25.8    &    5.6   &    21.6   & 40.1 \\
    \textcolor{red}{\cmark} &     \textbf{33.4}  &  \textbf{63.7}    &   \textbf{31.8}    &    \textbf{6.5}   &    \textbf{26.4}   & \textbf{44.1} \\
    \hline
    \\[-0.9em]  
    \multicolumn{7}{l}{MS COCO \textit{val} images:} \\
      \xmark    &   23.5    &       47.9 &      20.3 &  10.4     &      24.9 & 36.5 \\
    \textcolor{red}{\cmark} &   \textbf{26.0}    &  \textbf{50.0}     & \textbf{23.9}      & \textbf{10.8}      &   \textbf{28.5}    & \textbf{40.3} \\
    % \\[-0.9em]
    \Xhline{1pt}
    \end{tabular}%
    \vspace{-1.2em}
  \label{tab:mcg}%
\end{table}%

The concurrent method, Box2Seg~\cite{kulharia12356box2seg}, achieved an mIoU of 76.4\% on the PASCAL VOC validation images, but it is based on UperNet~\cite{xiao2018unified}, which is a more powerful segmentation network than DeepLab-v2~\cite{pytorchdeeplab}.
For a fair comparison between Box2Seg~\cite{kulharia12356box2seg} and our BBAM, we attempt to relieve the benefit of UperNet~\cite{xiao2018unified} over DeepLab-v2~\cite{chen2017deeplab} by comparing the relative performance of the weakly supervised model to the fully supervised model.
Box2Seg achieves 88.4\% of the performance of its fully supervised equivalent (76.4 \textit{vs.} 86.4); but the corresponding figure for BBAM and its fully supervised equivalent is 96.7\% (73.7 \textit{vs.} 76.2).

\begin{table*}[t]
  \centering
  %   Table1
\begin{minipage}{0.28\linewidth}
  \centering
  \includegraphics[width=\linewidth]{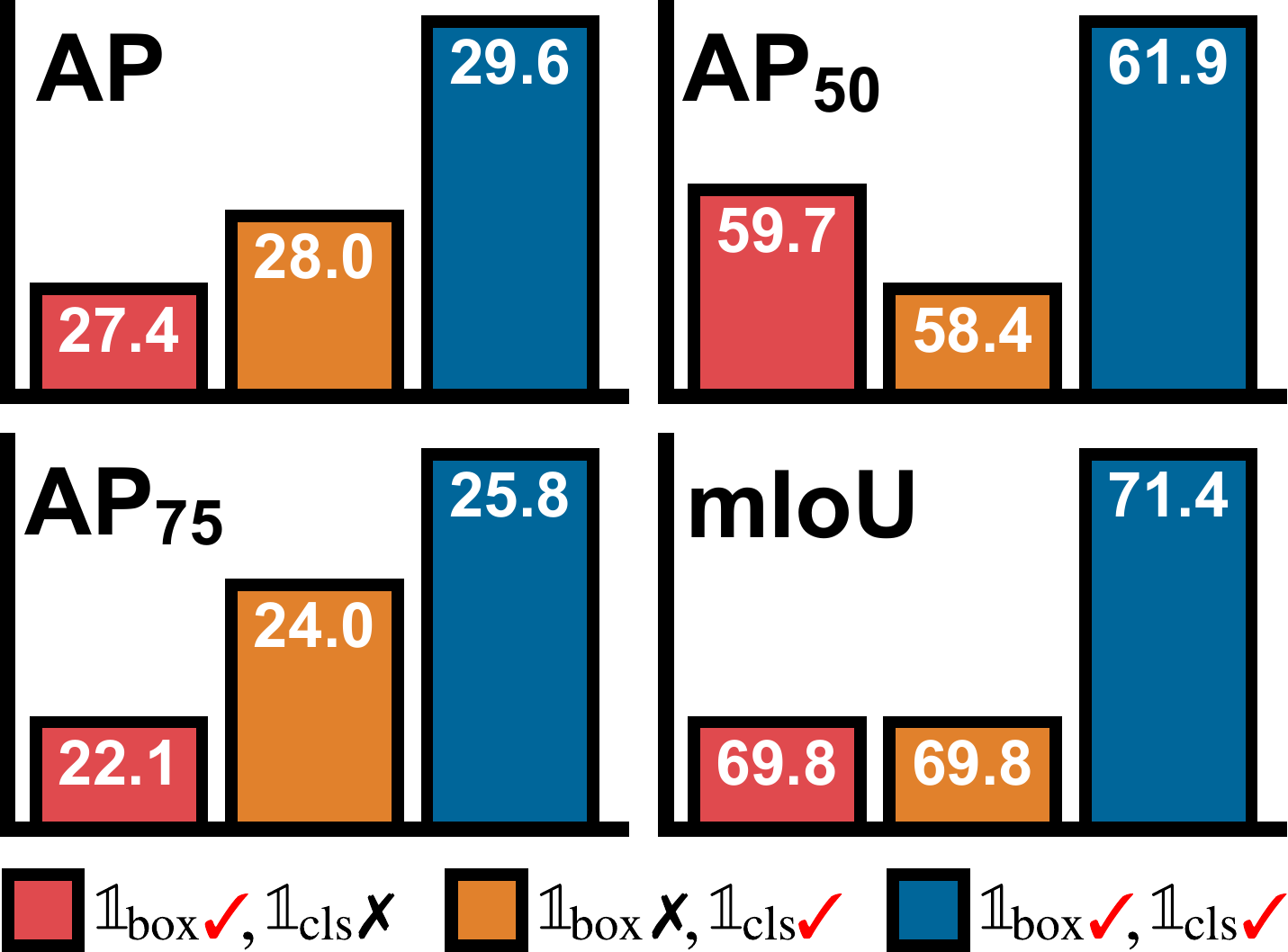}
  \vspace{-1.7em}
  \captionof{figure}{\label{fig_eachhead} Effect of each head on instance and semantic segmentation.}
  \end{minipage}\hfill
\begin{minipage}{0.34\linewidth}
\center
{
\begin{tabular}{c@{\hskip 0.15in}c@{\hskip 0.15in}c@{\hskip 0.2in}c@{\hskip 0.13in}c@{\hskip 0.09in}c}
     \Xhline{1pt}\\[-0.95em]
    $\theta_{\text{fg}}$    & $\theta_{\text{bg}}$ & $\mathcal{G}$ & AP   & AP$_{50}$  & AP$_{75}$ \\
    \hline\hline
    \\[-0.9em]
    0.2&0.2      &   \xmark    & 24.8      &  58.3     & 18.1  \\
    0.5&0.5  &   \xmark    &  28.3     &  59.5     &  24.7 \\

  0.8&0.8  &   \xmark    &  27.8     & 59.0      &  23.3 \\
    0.3   & 0.7   &    \xmark   &    28.1   &  59.5     & 24.0 \\
    0.3   & 0.7   &   \textcolor{red}{\cmark}    &  28.4   &    59.6  &  24.6  \\
    0.2   & 0.8   &    \xmark   &   28.6    &   60.4    & 24.0 \\
    0.2   & 0.8   &   \textcolor{red}{\cmark}    &  \textbf{29.6}     &  \textbf{61.9}     & \textbf{25.8} \\
    \Xhline{1pt}
    \end{tabular}%
    }
    \vspace{-0.7em}
      \caption{Analysis of thresholds $\theta_{\text{fg}}$ and $\theta_{\text{bg}}$, and effect of the growing technique $\mathcal{G}$.}
  \label{table_params}%
  \end{minipage}\hfill
%   Table3
\begin{minipage}{0.32\linewidth}
\centering
% \vspace{-0.25em}
{
    \begin{tabular}{c@{\hskip 0.17in}c@{\hskip 0.13in}c@{\hskip 0.1in}c@{\hskip 0.1in}c}
    \Xhline{1pt}\\[-0.95em]
               & \multicolumn{3}{c}{\texttt{Ins.}} & \texttt{Sem.} \\
    
    $\lambda$ & AP    & AP$_{50}$  & AP$_{75}$  & mIoU \\
    \hline\hline
    \\[-0.9em]
         0.001   &  26.6    &  58.7    &    21.1  &  67.9 \\
     0.003  &   28.1   &    59.9  & 22.8    &  69.7\\
     0.005    &   28.7   &   60.2    &   24.3    &  70.8 \\
     0.007   &  \textbf{29.6}     &      \textbf{61.9} &  \textbf{25.8}     &  \textbf{71.4}\\
     0.010   &  28.7    & 60.4     &     24.4  &  70.7 \\ 
     0.020   &   28.3   &    59.6  &     23.7  & 70.3 \\

    \Xhline{1pt}
    \end{tabular}%
    }
    \vspace{-0.7em}
      \caption{Effect of $\lambda$ on instance (\texttt{Ins.}) and semantic (\texttt{Sem.}) segmentation.}\label{table_lambda}%
\end{minipage}
\vspace{-1.2em}
\end{table*}

\vspace{-0.1em}
\subsection{Ablation Study}
\vspace{-0.1em}

\textbf{MCG proposals.}
Table~\ref{tab:mcg} shows how mask refinement with MCG proposals improves the instance segmentation performance of our method on the PASCAL VOC and MS COCO datasets.
Mask refinement with MCG proposals is particularly effective on masks for medium and large objects. 
The results obtained without MCG proposals offer the possibility of a fairer comparison with Hsu \textit{et al.}~\cite{hsu2019weakly}, which do not use MCG proposals. Our method produces better results than that of Hsu \textit{et al.}~\cite{hsu2019weakly} for both the PASCAL VOC and MS COCO datasets, which are shown in Tables~\ref{instance_voc} and \ref{instance_coco} respectively.
Hereinafter, to observe the contribution of each component of our system, we report results without using MCG proposals.

\textbf{\textit{Box} and \textit{cls heads}.}
BBAM can provide a separate attribution map for each head of the object detector by controlling the logical variables $\mathbb{1}_{\text{box}}$ and $\mathbb{1}_{\text{cls}}$ in Eq.~\ref{eq_mask_weak}.
Figure~\ref{fig_eachhead} shows the effect of the BBAM obtained from each head on the performance of weakly supervised semantic and instance segmentation. Using the BBAM obtained from either the \textit{box head} ($\mathbb{1}_{\text{box}}=1$ and $\mathbb{1}_{\text{cls}}=0$) or the \textit{cls head} ($\mathbb{1}_{\text{box}}=0$ and $\mathbb{1}_{\text{cls}}=1$) shows competent performance, but the best performance is achieved when the two heads are used together. 
We attribute this to the complementary property of the two heads, which is examined in more detail in Section~\ref{sec_analysis}.

\textbf{Parameter sensitivity analysis.} Table~\ref{table_params} shows the effect of the thresholds $\theta_{\text{fg}}$ and $\theta_{\text{bg}}$, and the seed growing technique $\mathcal{G}$.
When $\theta_{\text{fg}}$ equals to $\theta_{\text{bg}}$, all pixels are assigned to either the foreground or the background.
We see that ignoring some pixels can improve the AP values, and the seed growing technique further improves performance.
We then studied the effect of $\lambda$, which controls the sparsity of the BBAM, on the performance of weakly supervised semantic and instance segmentation, with the results shown in Table~\ref{table_lambda}. 
Our method shows similar performance on semantic and instance segmentation over a broad range of values of $\lambda$.

% \vspace{-0.4em}
\section{Detailed Analysis of the BBAM}\label{sec_analysis}
% \vspace{-0.3em}

\begin{figure}[t]
  \centering
%   \vspace{-1em}
  \includegraphics[width=\linewidth]{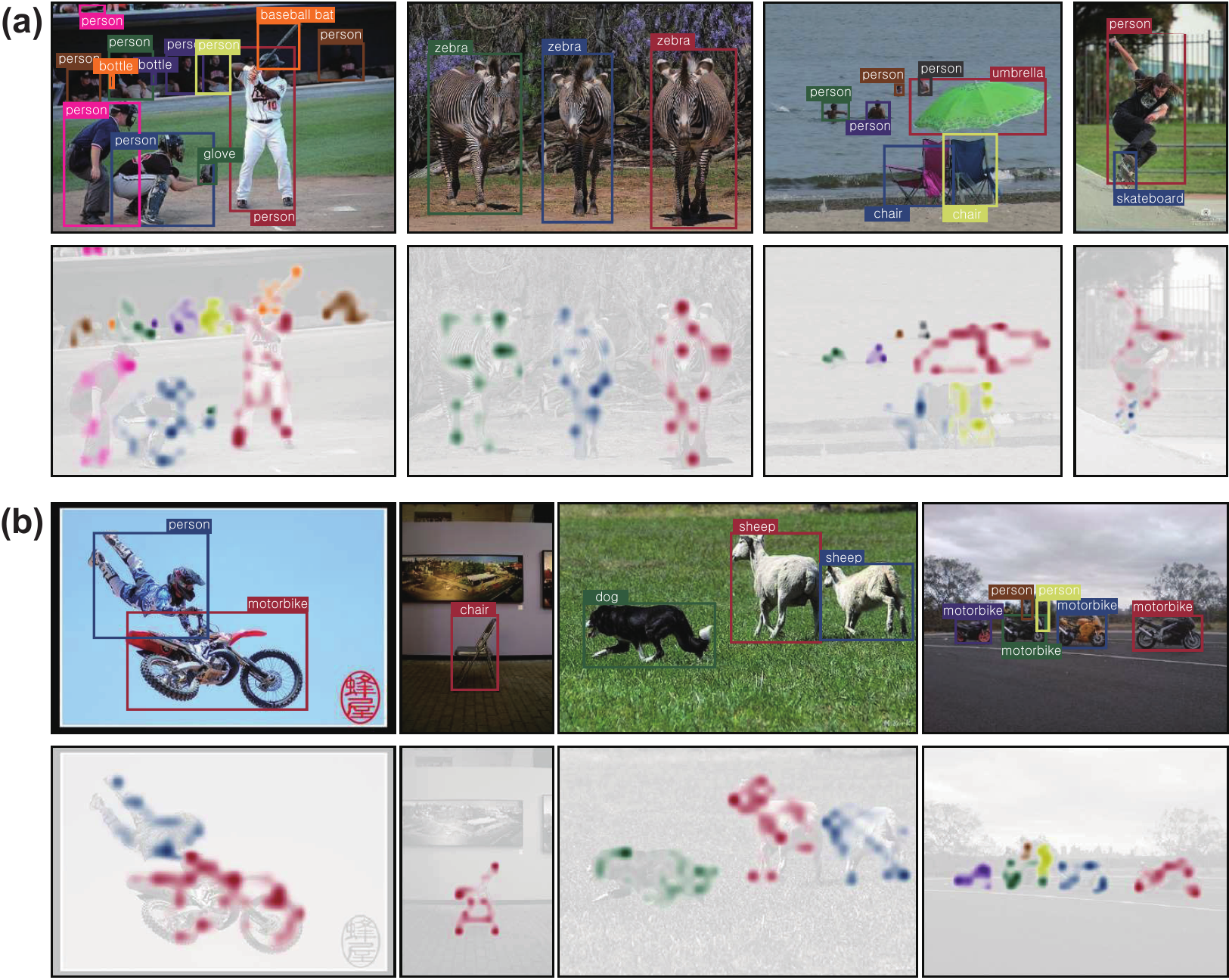} \\[-0.5em]
  \caption{\label{BBAM_samples_total} Examples of the predicted boxes and corresponding BBAMs. (a) BBAMs for MS COCO validation images. (b) BBAMs for PACSAL VOC validation images. Each BBAM corresponds to the predicted box of the same color.}
  \vspace{-1.2em}
\end{figure}

\textbf{Examples of BBAMs.}
Figure~\ref{BBAM_samples_total} shows BBAMs for validation images from PASCAL VOC~\cite{everingham2010pascal} and MS COCO~\cite{lin2014microsoft}.
The BBAMs have high values on the boundary and discriminative parts of each object, which are informative in conducting object detection.

\begin{figure*}[t]
  \centering
  \includegraphics[width=0.9\linewidth]{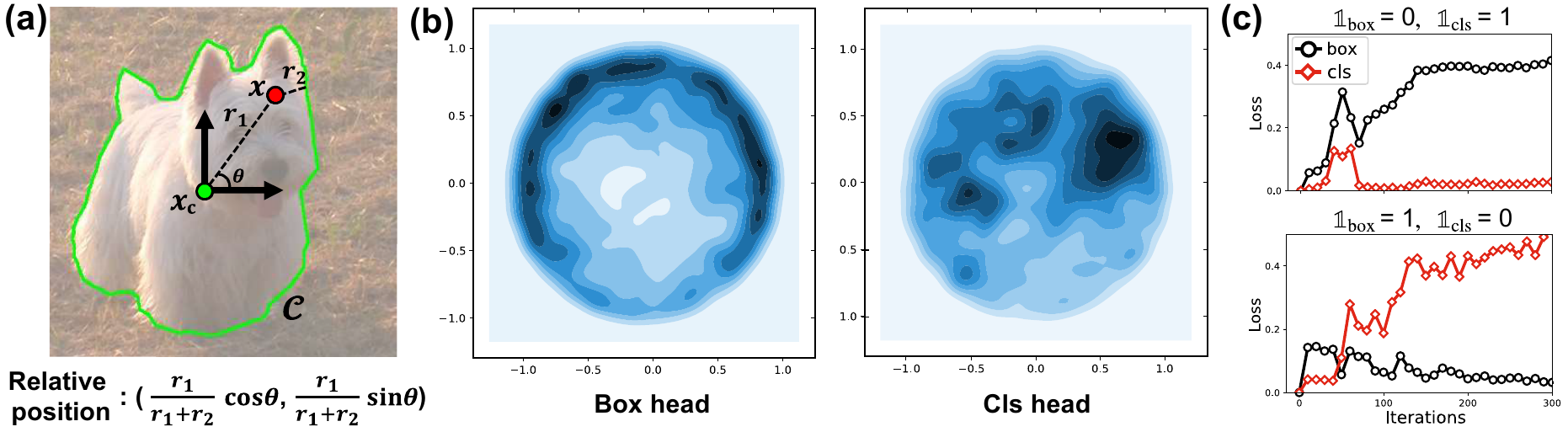} \\[-0.7em]
  \caption{\label{fig_complement} Complementary operation of the \textit{box head} and the \textit{cls head}. (a) The definition of relative position. (b) Relative positions of the highly activated pixels from each head. (c) \textit{Box} and \textit{class} loss curves.}
  \vspace{-1.2em}
\end{figure*}

\begin{figure}[t]
\centering
\includegraphics[width=\linewidth]{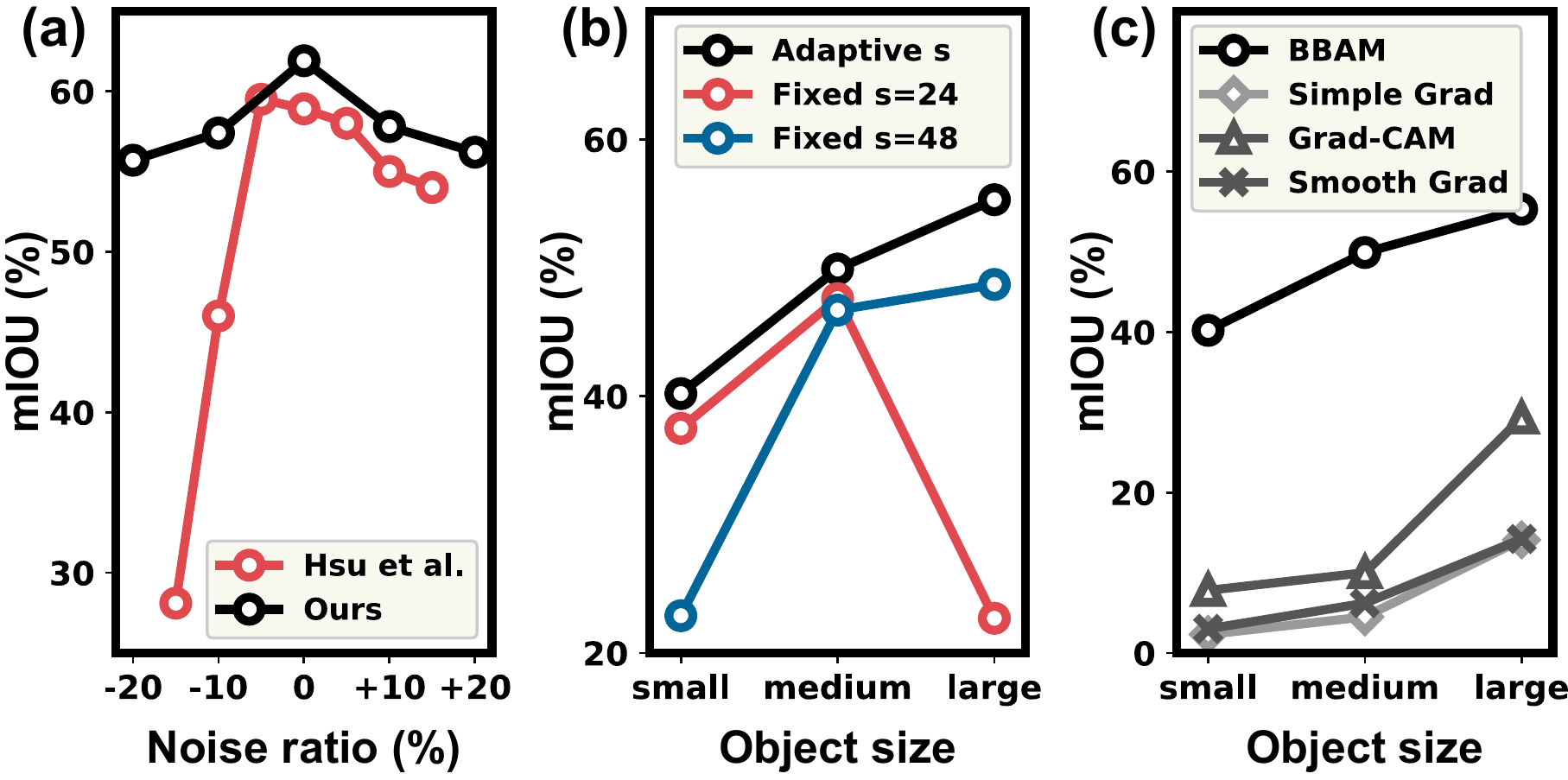}
\vspace{-1.5em}
\caption{\label{noise_robust} (a) Robustness against noisy box coordinate labels. (b) Localization accuracy by different strides. (c) Localization accuracy by different attribution methods.}
\vspace{-1em}
\end{figure}

\textbf{Complementary operation of the \textit{box} and \textit{cls heads}.}
To determine which regions of an object are important to each head, we investigated the distribution of high-value pixels in the BBAM produced by each head.
In Figure~\ref{fig_complement}(a), $\mathcal{C}$ is the set of points on the contour of the object mask, and $\vec{x_c}$ is its centroid.
For each pixel $\vec{x}$, we determine $r_1=\left\lVert \vec{x} - \vec{x_c} \right\rVert_2$ and $r_2 = \min_{\vec{c} \in \mathcal{C}} \left\lVert \vec{x} - \vec{c} \right\rVert_2$.
Letting the angle between $\vec{x}-\vec{x_c}$ and the $x$-axis be $\theta$, the position of the pixel $\vec{x}$ relative to $\vec{x_c}$ is $\vec{R}=(\frac{r_1}{r_1+r_2}\cos\theta, \frac{r_1}{r_1+r_2}\sin\theta)$.
In Figure~\ref{fig_complement}(b), we plot the relative positions of all the pixels with attribution values above 0.9 obtained from validation images of the PASCAL VOC dataset.
Pixels for which $\lVert \vec{R} \rVert_2 \approx 1$ are near the boundary of the object.
We observed that high values attributed by the \textit{box head} mainly occur near the boundary of the object, and those by the \textit{cls head} mainly occur in the interior.

Furthermore, we observed how much the prediction of each head changes when either of $\mathbb{1}_{\text{box}}$ and $\mathbb{1}_{\text{cls}}$ is set to 1 during the optimization of Eq.~\ref{eq_mask}.
The extent of the change in prediction of each head can be inferred from the corresponding loss in Eq.~\ref{eq_mask_perturb}. Figure~\ref{fig_complement}(c) shows that applying the optimization of Eq.~\ref{eq_mask} to one of the heads increases the loss of the other head, implying that the discriminative area of the image necessary for each head is not sufficient for the other head to maintain the prediction.
These two observations suggest that the BBAM of each head provides complementary attributions. Examples of BBAMs obtained from each head are presented in the Appendix.

\textbf{Label noise in object detection.} We also looked at the robustness of our system against noisy box coordinate labels in instance segmentation. 
Hsu \textit{et al.}~\cite{hsu2019weakly} considered the effect of up to $\pm15\%$ of label noise: we extend this to $\pm20\%$. 
The validity of the bounding box tightness priors used by Hsu \textit{et al.}~\cite{hsu2019weakly} is seriously compromised by inaccurate box coordinates, with a considerable effect on performance, as shown in Figure~\ref{noise_robust}(a).
Our method shows better robustness than that of Hsu \textit{et al.}~\cite{hsu2019weakly}, whether the noise consists of expanded or contracted bounding box annotations.

\textbf{Effectiveness of an adaptive stride $\boldsymbol{s(a)}$.}
As mentioned in Section~\ref{method_BBAM}, we use an adaptive stride $16 \leq s(a) \leq 64$ to cope with feature transformation due to RoI pooling.
Figure~\ref{noise_robust}(b) shows the IoU between the BBAM and ground truth mask on PASCAL VOC validation images, along with the results using fixed strides of 24 and 48.
Figure~\ref{noise_robust}(b) shows that a small fixed stride ($s$=24) is ineffective with large objects, as is a large fixed stride ($s$=48) with small objects. By contrast, an adaptive stride $s(a)$ can deal with objects of various sizes.

\begin{figure}[t]
\centering
\includegraphics[width=\linewidth]{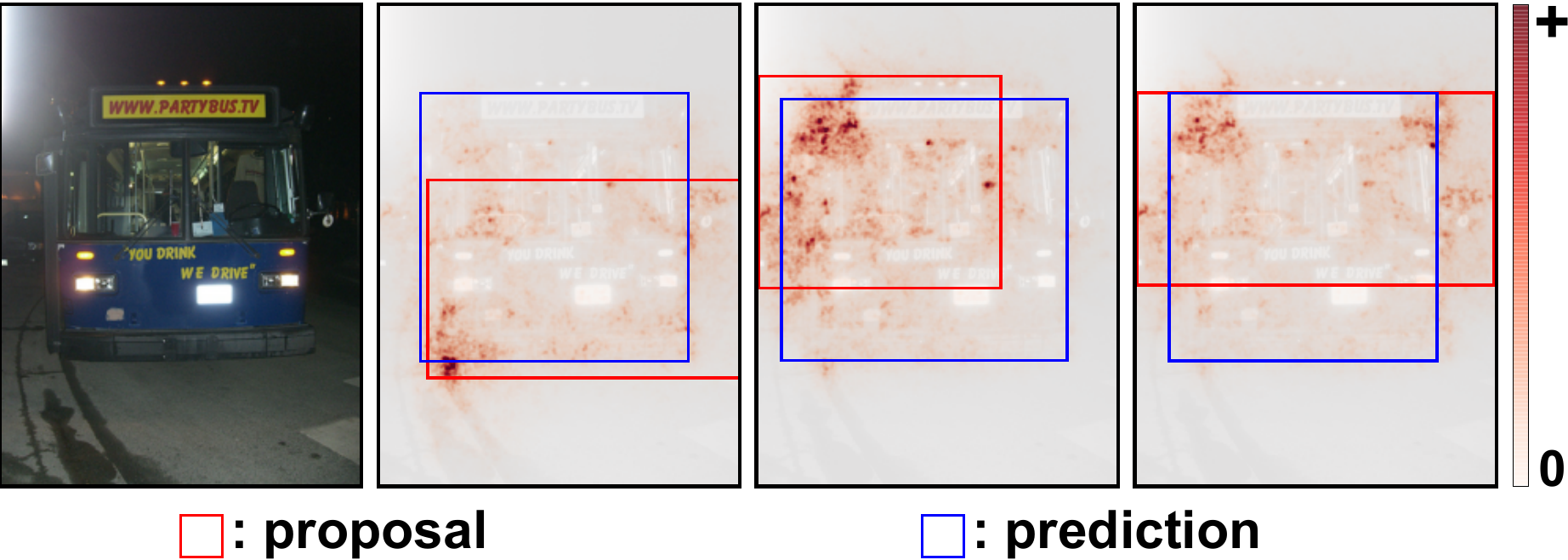}
\vspace{-1.5em}
\caption{\label{gradient_ex} Examples of SimpleGrad~\cite{zeiler2014visualizing} for three similar predictions obtained from different proposals.}
\vspace{-1em}
\end{figure}

\textbf{Comparison with gradient-based methods.}
Gradient-based attribution methods, such as SimpleGrad~\cite{zeiler2014visualizing}, SmoothGrad~\cite{smilkov2017smoothgrad}, and Grad-CAM~\cite{selvaraju2017grad} can also provide attributions for the output of an object detector. 
However, since only the subset of features associated with the imperfect proposal is delivered to the \textit{cls} and \textit{box} heads, the gradients with respect to pixels, which exist outside the proposal yet essential for prediction, can vanish (but not completely, due to the receptive field). 
We provide empirical results supporting this analysis on the PASCAL VOC validation images: \underline{\textbf{(1)}} Figure~\ref{gradient_ex} shows examples in which SimpleGrad~\cite{zeiler2014visualizing} is applied to three similar predictions from different proposals. Pixels outside the proposal do indeed influence the predictions, but SimpleGrad's attributions mainly appear inside the proposal. 
\underline{\textbf{(2)}} We observed that the majority (87\%) of pixels with attribution values above 0.9 appear inside the imperfect proposal; the mean IoU between the set of positive proposals and the corresponding predictions is low (\textit{i.e.,} 0.56).
\underline{\textbf{(3)}} Figure~\ref{noise_robust}(c) shows that attribution maps from gradient-based attribution methods correlate poorly with ground truth masks.

\vspace{-0.15em}
\section{Conclusions}
\vspace{-0.15em}
We have introduced a bounding box attribution map (BBAM), which provides pixel-level localization of each target object in its bounding box by finding the smallest region that preserves the predictions of the object detector.
Our formulation is built on two-stage object detectors, but applying our method to one-stage object detectors is straightforward as long as they have \textit{box} and \textit{cls} heads.
Our experiments demonstrate that the BBAM achieves state-of-the-art performance on the PASCAL VOC and MS COCO benchmarks in weakly supervised semantic and instance segmentation.
We have also analyzed BBAMs from various viewpoints, and compared our technique with other attribution methods, to provide a deeper understanding of our approach.
We expect BBAMs to be a staple of future work on weakly supervised semantic and instance segmentation with bounding boxes, on a par with the CAM for class labels.

\bigskip
\noindent\textbf{Acknowledgements:}
This work was supported by the National Research Foundation of Korea (NRF) grant funded by the Korea government (MSIT) [2018R1A2B3001628], AIR Lab (AI Research Lab) in Hyundai \& Kia Motor Company through HKMC-SNU AI Consortium Fund, and the Brain Korea 21 Plus Project in 2021.

{\small
% \balance
\bibliographystyle{ieee_fullname}
\bibliography{egbib}
}

\setcounter{section}{0}
\renewcommand\thesection{\Alph{section}}
\setcounter{table}{0}
\renewcommand{\thetable}{A\arabic{table}}
\setcounter{figure}{0}
\renewcommand{\thefigure}{A\arabic{figure}}

\clearpage

\renewcommand{\tabcolsep}{2pt}

\begin{table*}[t]
  \caption{Comparison of per-class mIoU scores.}
  \vspace{-1em}
% \resizebox{\textwidth}{!}{%
  \centering
  \begin{adjustbox}{max width=\textwidth}
    \begin{tabular}{lccccccccccccccccccccc|c}
    
        \Xhline{1pt}

        \\[-0.95em]
     & bkg& aero  & bike  & bird  & boat  & bottle & bus   & car   & cat   & chair & cow   & table & dog   & horse & motor & person & plant & sheep & sofa  & train & tv  ~ & mIOU \\
   
    \hline
    \hline
    \\[-0.9em]

   \multicolumn{22}{l}{Results on validation images:}\\

    Shen \textit{et al.}~\cite{shen2018bootstrapping}~~~~~~&   86.8    &   71.2    &   32.4    &   77.0    &   24.4   &   69.8    &    85.3  &    71.9   &    86.5   &  27.6     &   78.9    &   40.7    &    78.5 &    79.1   &   72.7    &  73.1 &  49.6   &74.8  & 36.1  &  48.1 &   59.2~ &  63.0\\
    CIAN~\cite{fan2018cian}&   88.2    &   79.5    &   32.6    &   75.7    &   56.8   &   72.1    &    85.3  &    72.9   &    81.7   &  27.6     &   73.3    &   39.8    &    76.4 &    77.0   &   74.9    &  66.8 &  46.6   &    81.0  & 29.1  &  60.4 &   53.3~ &  64.3\\
    FickleNet~\cite{lee2019ficklenet} &   89.5    &   76.6    &   32.6    &   74.6    &   51.5   &   71.1    &    83.4  &    74.4   &    83.6   &  24.1     &   73.4    &   47.4    &    78.2 &    74.0   &   68.8    &  73.2 &  47.8   &    79.9  & 37.0  &  57.3 &   64.6~ &  64.9\\
    SSDD~\cite{Shimoda_2019_ICCV} &   89.0    &   62.5    &   28.9    &   83.7    &   52.9   &   59.5    &    77.6  &    73.7   &    87.0   &  34.0     &   83.7    &   47.6    &    84.1 &    77.0   &   73.9    &  69.6 &  29.8   &    84.0  & 43.2  &  68.0 &   53.4~ &  64.9\\
    Lee \textit{et al.}~\cite{lee2019frame} &   90.8    &   82.2    &   35.1    &  82.4    &   72.2   &   71.4    &    82.7  &    75.0   &    86.9   &  18.3     &   74.2    &   29.6    &    81.1&    79.2   &   74.7    &  76.4 &  44.2   &    78.6  & 35.4  &  72.8 &   63.0~ &  66.5\\

    BBAM (Ours) &   92.7    &   80.6    &   33.8    &  83.7    &   64.9   &   75.5    &    91.3  &    80.4   &    88.3   &  37.0     &   83.3    &   62.5    &    84.6 &    80.8   &   74.7    &  80.0 &  61.6   &    84.5  & 48.6  &  85.8 &   71.8~ &  73.7\\
    \hline\\[-0.9em]
    \multicolumn{22}{l}{Results on test images:}\\
    Shen \textit{et al.}~\cite{shen2018bootstrapping}&   87.2    &   76.8    &   31.6    &   72.9   &   19.1   &   64.9    &    86.7  &    75.4   &    86.8   &  30.0     &   76.6    &   48.5    &    80.5 &    79.9   &   79.7    &  72.6 &  50.1   &83.5  & 48.3  &  39.6 &   52.2~ &  63.9\\
    FickleNet~\cite{lee2019ficklenet} &   90.3    &   77.0    &   35.2    &   76.0    &   54.2   &   64.3    &    76.6  &    76.1   &    80.2   &  25.7     &   68.6    &   50.2    &    74.6&    71.8   &   78.3    &  69.5 &  53.8   &    76.5  & 41.8  &  70.0 &   54.2~ &  65.0\\
    SSDD~\cite{Shimoda_2019_ICCV} &   89.0    &   62.5    &   28.9    &   83.7    &   52.9   &   59.5    &    77.6  &    73.7   &    87.0   &  34.0     &   83.7    &   47.6    &    84.1 &    77.0   &   73.9    &  69.6 &  29.8   &    84.0  & 43.2  &  68.0 &   53.4~ &  64.9\\
    Lee \textit{et al.}~\cite{lee2019frame} &   91.2    &   84.2    &   37.9    &   81.6    &   53.8   &   70.6    &    79.2  &    75.6   &    82.3   &  29.3     &   76.2    &   35.6    &    81.4&    80.5   &   79.9    &  76.8 &  44.7   &    83.0  & 36.1  &  74.1 &   60.3  &  67.4\\
    BBAM (Ours)&   92.8    &   83.5    &   33.4    &   88.9    &   61.8   &   72.8    &    90.3  &   83.5   &    87.6   &  34.7     &   82.9    &   66.1    &    83.9 &  81.1 & 78.3   &   77.4    &  55.2 &  86.7   &    58.5  & 81.5  &  66.4 &   73.7\\
        \Xhline{1pt}
    \vspace{-2em}
    \end{tabular}%
  \end{adjustbox}%
  \label{class-specific-results}%
\end{table*}%

\section{Appendix}
\subsection{Implementation details}
\textbf{TV norm.} To suppress the artifacts in the mask $\mathcal{M}$, we regularized $\mathcal{M}$ with total variation (TV) norm in Eq. \textcolor{red}{1} in the main paper, as done in Fong \textit{et al.}~\cite{fong2017interpretable}. The resulting loss function to find the best $\mathcal{M}^{*}$ becomes:
\begin{align}\label{eq_mask_supp}
\begin{split}
\mathcal{L}_{\mathcal{M}} = &\lambda \left\lVert \mathcal{M} \right\rVert_1 + \lambda_{\text{TV}} \left\lVert  \nabla \mathcal{M} \right\rVert_{\beta}^{\beta} \\ &+ \mathbb{1}_{\text{box}} \left\lVert t^c - f^{\text{box}}(\Phi(I, \mathcal{M}), o) \right\rVert_1 \\ &+ \mathbb{1}_{\text{cls}} \left\lVert p^c - f^{\text{cls}}(\Phi(I, \mathcal{M}), o) \right\rVert_1,
\end{split}
\end{align}
where $\lambda_{\text{TV}}$ is a balancing factor for TV norm. We set $\lambda_{\text{TV}}$ to $10^{-4}$ and $\beta$ to $3$.
We observed that the resulting mask $\mathcal{M}^{*}$ has a little dependency on the value of $\lambda_{\text{TV}}$.

We can find the best $\mathcal{M}^{*}$ by using gradient descent with respect to $\mathcal{M}$. Letting the mask at iteration $t$ be $\mathcal{M}^{t}$, the mask at iteration $t+1$ can be expressed as
\begin{align}\label{sgd}
\mathcal{M}^{t+1} = \mathcal{M}^{t} - \xi \nabla_{\mathcal{M}^{t}} \mathcal{L}_{\mathcal{M}^{t}},
\vspace{-0.5em}
\end{align}
where $\xi$ is a learning rate. Indeed, the update in Eq~\ref{sgd} was implemented through Adam optimizer.

\textbf{Optimization details for semantic segmentation.}
We used the default setting provided by~\cite{pytorchdeeplab}, except for the batch size, the number of training iterations, and the learning rate. We set the batch size to 8, the number of training iterations to $2.4\times10^4$, and the learning rate to $2\times10^{-4}$.

\textbf{Optimization details for instance segmentation on the PASCAL VOC dataset.}
Regarding the characteristics of the PASCAL VOC dataset~\cite{everingham2010pascal}, we adjusted the input image size and the anchor size accordingly. We set the max and min size of training images to 800 and 512, respectively, and anchor sizes for each FPN level to [21, 42, 84, 168, 332]. We trained Mask R-CNN~\cite{he2017mask} with a learning rate $8\times10^{-3}$ for $2\times10^4$ iterations.

\textbf{Optimization details for instance segmentation on the MS COCO 2017 dataset.}
We followed the default settings provided by maskrcnn-benchmark repository~\cite{massa2018mrcnn}.

\textbf{Post-processing of semantic and instance segmentation.}
CRF~\cite{krahenbuhl2011efficient} is a popular post-processing technique for semantic and instance segmentation~\cite{hsu2019weakly, lee2019ficklenet, lee2019frame, huang2018weakly, Shimoda_2019_ICCV}. 
We also used CRFs as a post-processing method for semantic and instance segmentation.

\subsection{Additional Results}

\textbf{Comparison of per-class mIoU scores.}
Table~\ref{class-specific-results} shows the per-class mIoU of our method and recently produced methods.

\begin{figure*}[t]
  \centering
  \vspace{-0.2em}
  \includegraphics[width=\linewidth]{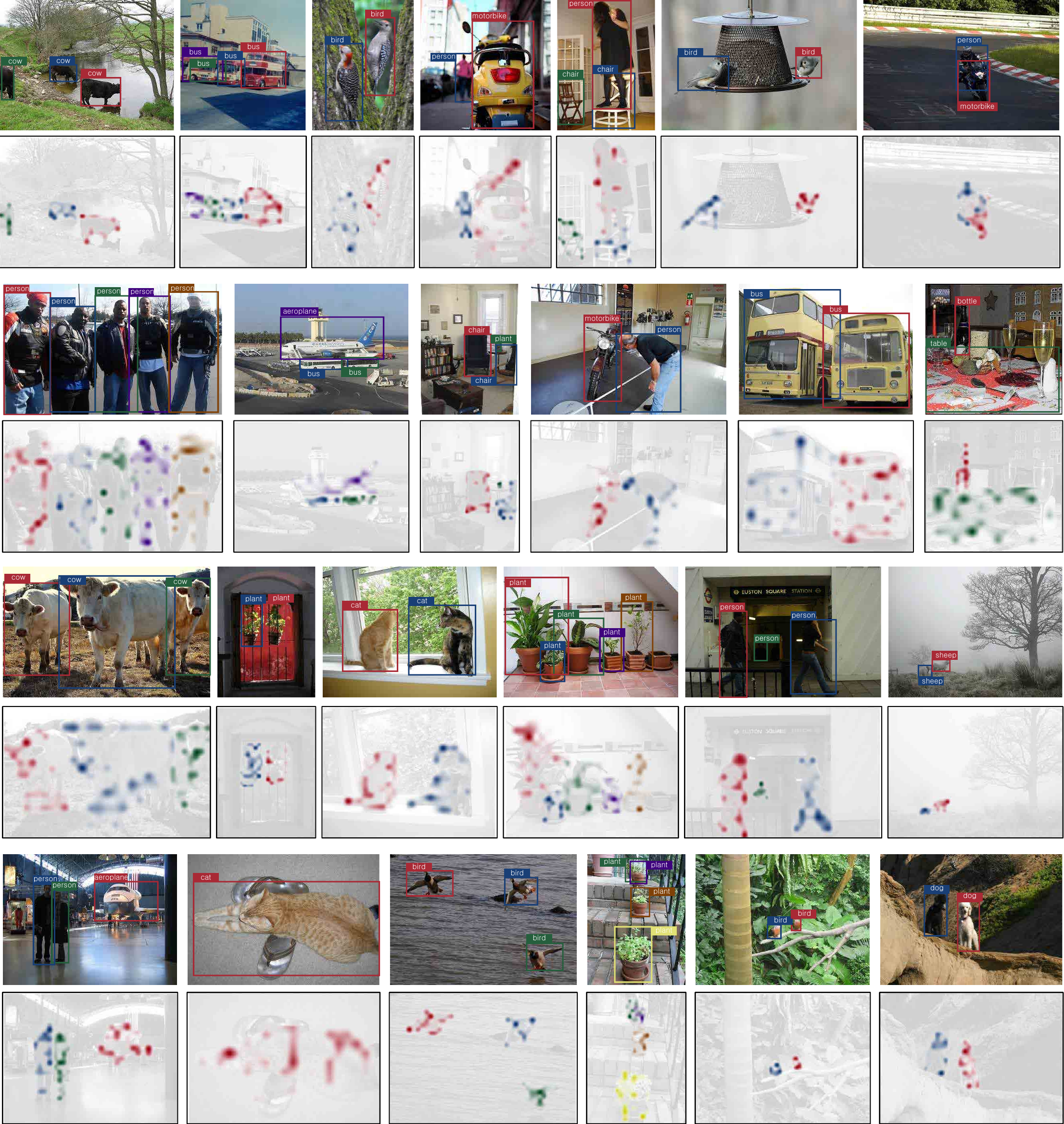} \\[-0.5em]
  \caption{\label{BBAM_samples_faster} Examples of PASCAL VOC~\cite{everingham2010pascal} validation images with the results of object detection and corresponding BBAMs, obtained from Faster R-CNN~\cite{ren2015faster}.}
\end{figure*}

\begin{figure*}[t]
  \centering
  \vspace{-0.2em}
  \includegraphics[width=\linewidth]{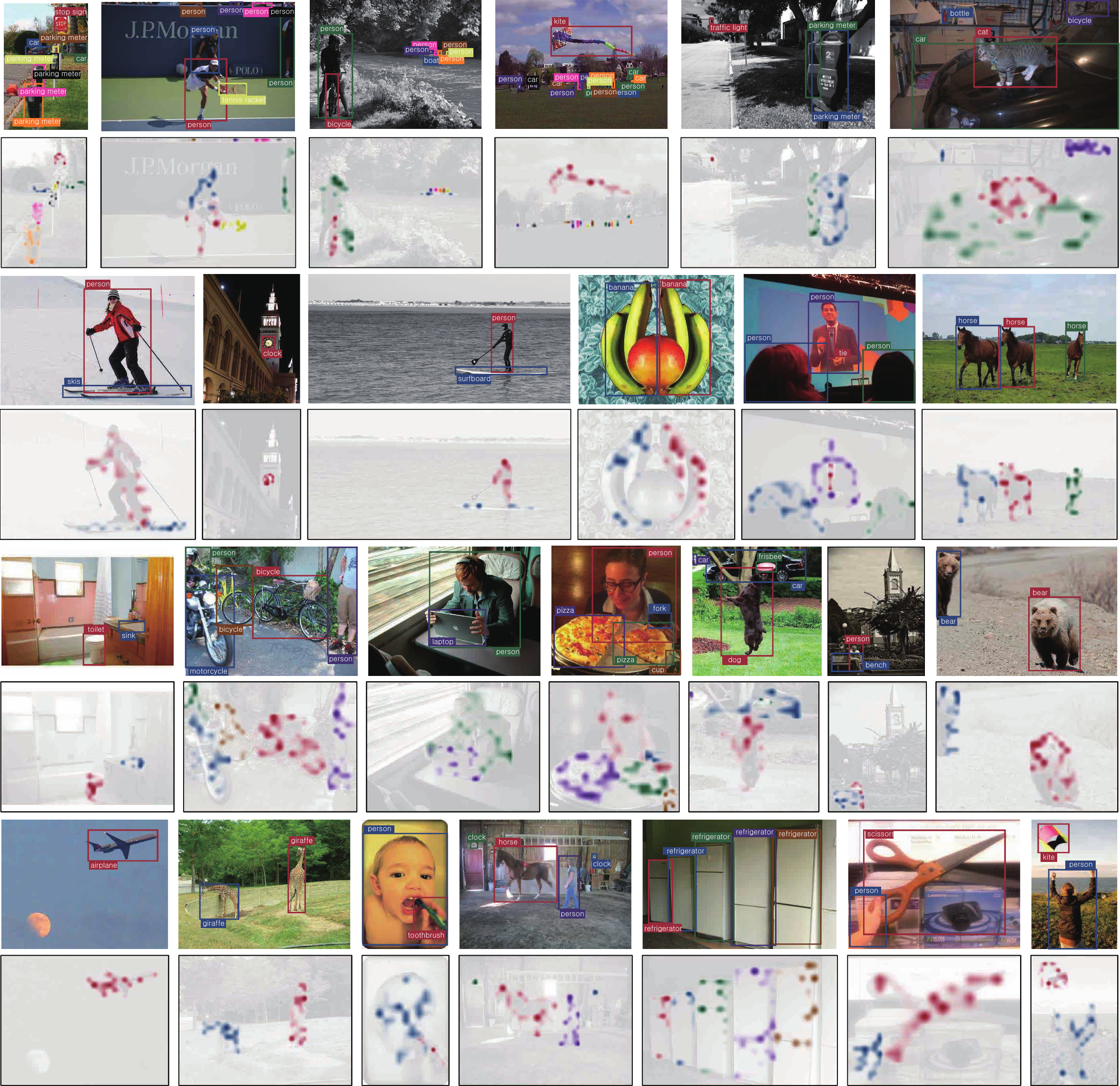} \\[-0.5em]
  \caption{\label{BBAM_samples_coco} Examples of MS COCO 2017~\cite{lin2014microsoft} validation images with the results of object detection and corresponding BBAMs, obtained from Faster R-CNN~\cite{ren2015faster}.}
\end{figure*}

\textbf{More examples of BBAMs.}
We present more examples of BBAMs for PASCAL VOC~\cite{everingham2010pascal} validation images with Faster R-CNN~\cite{ren2015faster} (Figure~\ref{BBAM_samples_faster}) and for MS COCO 2017~\cite{lin2014microsoft} validation images with Faster R-CNN~\cite{ren2015faster} (Figure~\ref{BBAM_samples_coco}).

\begin{figure*}[t]
  \centering
  \vspace{-0.2em}
  \includegraphics[width=\linewidth]{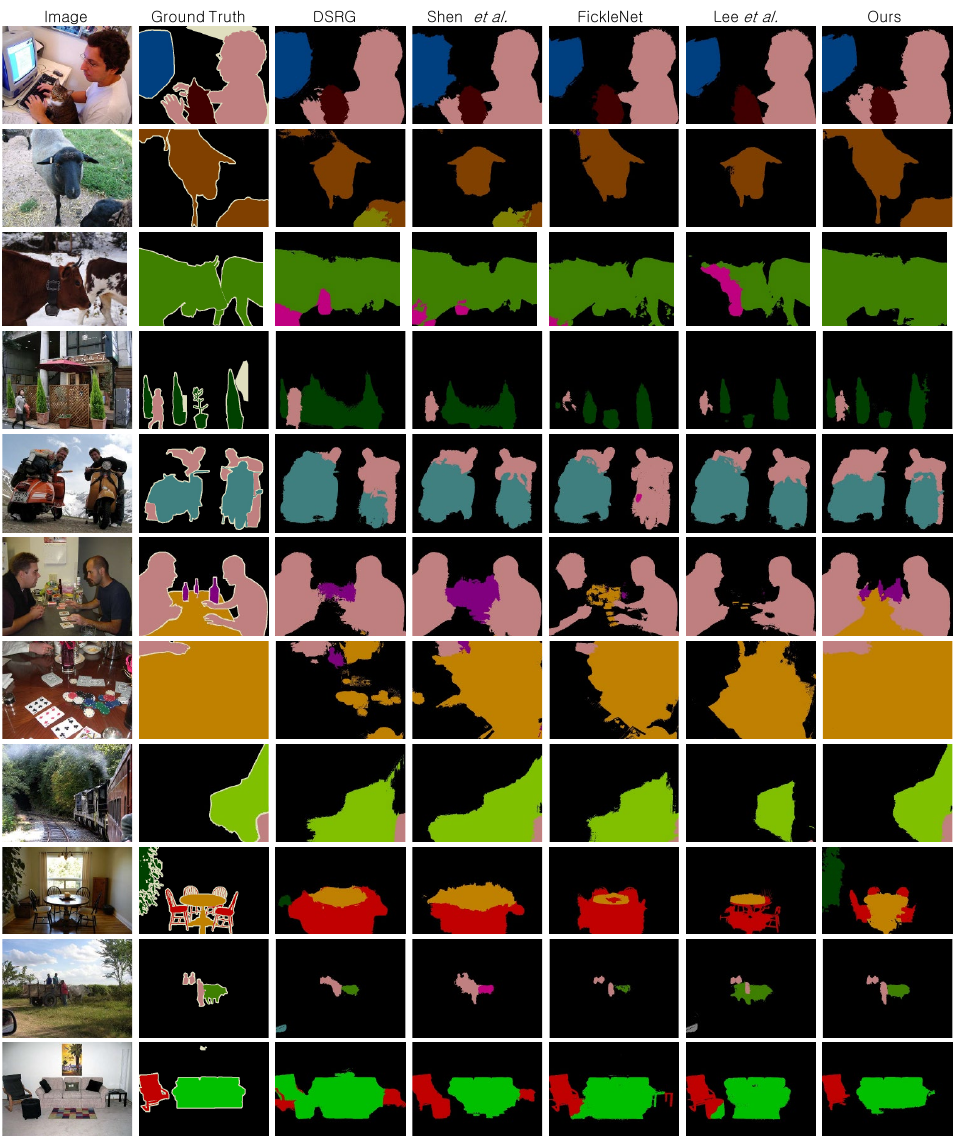} \\[-0.5em]
  \caption{\label{seg_ex} Examples of predicted semantic masks for PASCAL VOC validation images of DSRG~\cite{huang2018weakly}, Shen \textit{et al.}~\cite{shen2018bootstrapping}, FickleNet~\cite{lee2019ficklenet}, Lee \textit{et al.}~\cite{lee2019frame}, and our method.}
\end{figure*}

\textbf{Additional mask examples on semantic segmentation.}
Figure~\ref{seg_ex} shows more examples of the semantic masks produced by DSRG~\cite{huang2018weakly}, Shen \textit{et al.}~\cite{shen2018bootstrapping}, FickleNet~\cite{lee2019ficklenet}, Lee \textit{et al.}~\cite{lee2019frame}, and our method.

\begin{figure*}[t]
  \centering
  \vspace{-0.2em}
  \includegraphics[width=\linewidth]{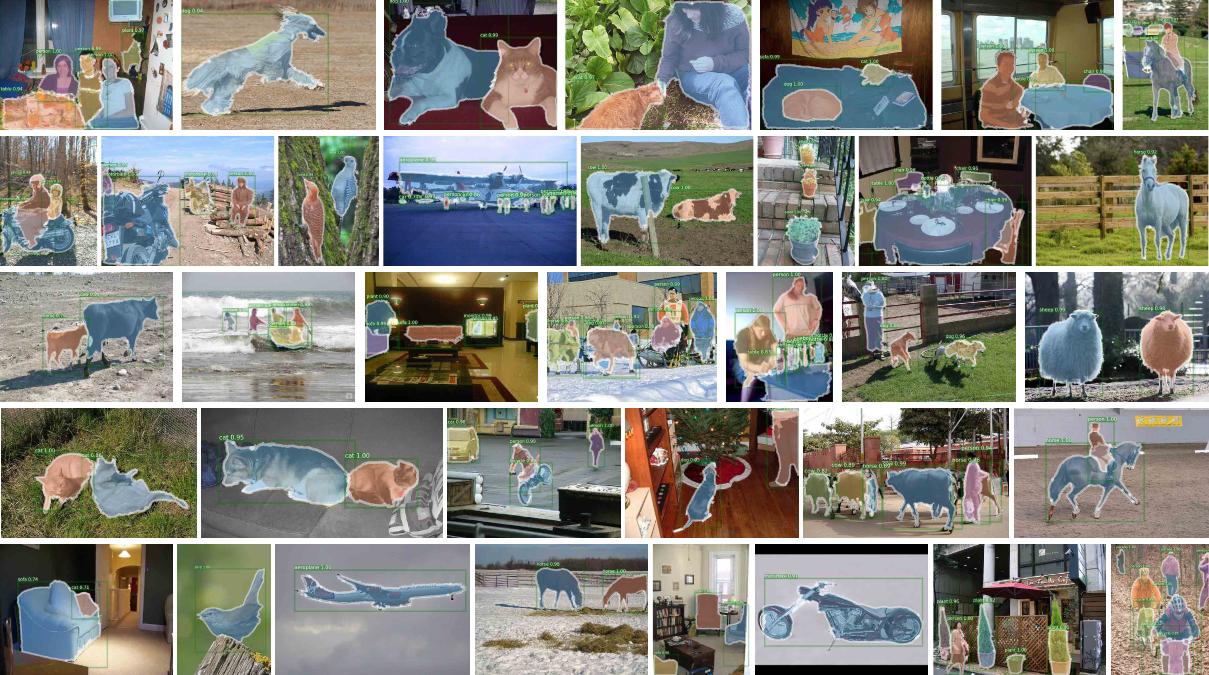} \\[-0.5em]
  \caption{\label{ins_seg_voc} Examples of predicted instance masks for PASCAL VOC validation images of our method.}
\end{figure*}

\begin{figure*}[t]
  \centering
  \vspace{-0.2em}
  \includegraphics[width=\linewidth]{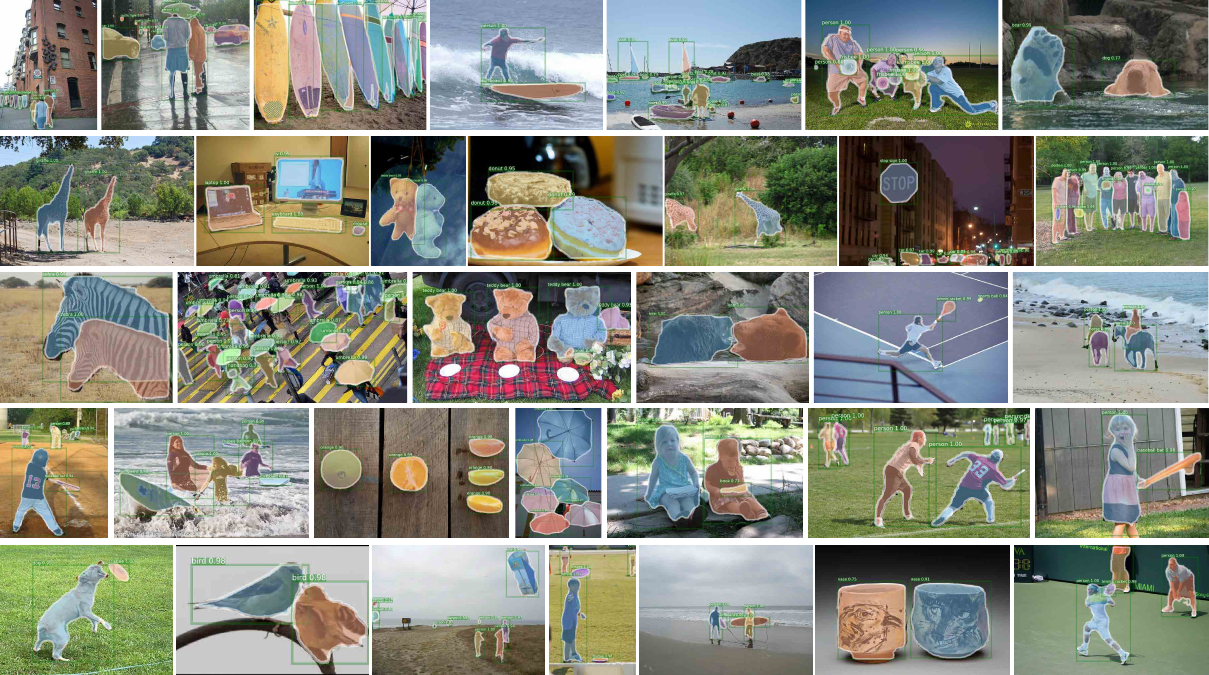} \\[-0.5em]
  \caption{\label{ins_seg_coco} Examples of predicted instance masks for MS COCO 2017 validation images of our method.}
\end{figure*}

\textbf{More mask examples on instance segmentation.}
Figure~\ref{ins_seg_voc} shows more examples of the instance masks on PASCAL VOC 2012 validation images obtained from IRNet~\cite{ahn2019weakly}, Hsu \textit{et al.}~\cite{hsu2019weakly}, and our method.
Figure~\ref{ins_seg_coco} shows examples of instance masks on MS COCO 2017 validation images obtained by our method.


\end{document}